\documentclass[11pt]{article}

\usepackage[final]{acl}

\usepackage{times}
\usepackage{latexsym}
\usepackage{natbib}
\usepackage[T1]{fontenc}

\usepackage[utf8]{inputenc}

\usepackage{microtype}

\usepackage{inconsolata}

\usepackage{graphicx}

\usepackage{microtype}
\usepackage{graphicx}
\usepackage{subcaption}
\usepackage{booktabs} 
\usepackage{xspace}
\usepackage{booktabs}
\usepackage{xcolor}
\usepackage{colortbl} 
\usepackage{times}
\usepackage{latexsym}
\usepackage[font={small}]{caption}
\usepackage{enumitem}
\usepackage{xspace}
\usepackage{hyperref}
\usepackage{booktabs}
\usepackage{multirow}
\usepackage{natbib}
\usepackage{xurl} 
\usepackage{listings}
\usepackage{xcolor}

\usepackage{tabularx}
\usepackage{xparse}
\usepackage{fvextra}  

\ifnum\pdfshellescape=1\relax
  \usepackage{minted}
  \setminted{
    fontsize=\footnotesize,
    breaklines=true,
    breakanywhere=true,
    linenos=false,
    frame=single,
    tabsize=2
  }
\else
  \NewDocumentEnvironment{minted}{O{} m}{%
    \VerbatimEnvironment
    \begin{Verbatim}[
      fontsize=\footnotesize,
      breaklines=true,
      breakanywhere=true,
      numbers=none,
      frame=single,
      tabsize=2,
      #1
    ]%
  }{%
    \end{Verbatim}%
  }

  \NewDocumentCommand{\mintinline}{O{} m m}{\texttt{#3}}

  \NewDocumentCommand{\setminted}{m}{}
  \NewDocumentCommand{\usemintedstyle}{m}{}
  \NewDocumentCommand{\inputminted}{O{} m m}{%
    \VerbatimInput[
      fontsize=\footnotesize,
      breaklines=true,
      breakanywhere=true,
      frame=single,
      tabsize=2,
      #1
    ]{#3}%
  }
\fi
\usepackage{hyperref}

\usepackage[most]{tcolorbox}
\usepackage{graphicx}
\usepackage{booktabs}
\usepackage{colortbl}
\usepackage{xcolor}
\newtcolorbox{promptbox}{
  breakable,
  colback=white,
  colframe=black!25,
  boxrule=0.4pt,
  arc=2pt,
  left=6pt,right=6pt,top=6pt,bottom=6pt
}

\usepackage{amsmath}
\usepackage{amssymb}
\usepackage{mathtools}
\usepackage{amsthm}
\usepackage{makecell}
\usepackage[capitalize,noabbrev]{cleveref}
\usepackage{xcolor}
\usepackage{listings}

\usepackage[most]{tcolorbox}
\tcbuselibrary{listingsutf8,breakable}

\newtcblisting{judgepromptbox}{
  listing only,
  breakable,
  colback=black!2,
  colframe=black!25,
  boxrule=0.3pt,
  arc=1.5pt,
  left=4pt,right=4pt,top=4pt,bottom=4pt,
  listing options={
    basicstyle=\ttfamily\footnotesize,
    columns=fullflexible,
    breaklines=true,
    breakatwhitespace=true,
    breakautoindent=true,
    keepspaces=true,
    showstringspaces=false,
    postbreak=\mbox{\textcolor{black!50}{\tiny$\hookrightarrow$}\space},
    numbers=none
  }
}

\theoremstyle{plain}

\theoremstyle{definition}

\theoremstyle{remark}

\usepackage[textsize=tiny]{todonotes}

\newcommand{\emphbf}[1]{{\textbf{#1}\xspace}}

\newcommand{\tname}[1]{\textsc{#1}\xspace}
\newcommand{\mypara}[1]{\smallskip\noindent\emphbf{#1.}\xspace}
\newcommand{\tool}{\tname{AgentRx}}
\newif\ifdraft
\draftfalse

\ifdraft
\newcommand\authorrnote[3]{\textcolor{#1}{({#2}: {#3})}\xspace}
\newcommand{\TODO}[1]{{\color{orange!80!black}[\textsl{#1}]}}

\else
\newcommand\authorrnote[2]{}
\newcommand{\TODO}[1]{}
\newcommand{\CHETAN}[1]{}

\fi
\title{\tool: Diagnosing AI Agent Failures from Execution Trajectories}

\author{Shraddha Barke \\
  Microsoft Research \\
  \texttt{sbarke@microsoft.com} \\\And
  Arnav Goyal\thanks{\hspace{0.2em}Work done during an internship at Microsoft.} \\  Microsoft \\
  \texttt{20arnav@gmail.com} \\\And
  Alind Khare \\
  Microsoft \\
  \texttt{alindkhare@microsoft.com} \\\AND
  Avaljot Singh\footnotemark[1] \\
  UIUC \\
  \texttt{avaljot2@illinois.edu} \\\And
  Suman Nath \\
  Microsoft \\
  \texttt{Suman.Nath@microsoft.com} \\\And
  Chetan Bansal \\
  Microsoft \\
  \texttt{chetanb@microsoft.com} \\}

\begin{document}



\maketitle
\begin{abstract}
AI agents often fail in ways that are difficult to localize because executions are probabilistic, long-horizon, multi-agent, and mediated by noisy tool outputs.
We address this gap by manually annotating failed agent runs and release a novel benchmark of 170 trajectories across 11 diverse task settings, including structured API workflows, incident management, and open-ended web/file tasks.
Each trajectory is annotated with a critical failure step and a category from a grounded-theory derived, cross-domain failure taxonomy.
To mitigate the human cost of failure attribution, we present \tool, an \emph{automated domain-agnostic} diagnostic framework that pinpoints the critical failure step in a failed agent trajectory.
It synthesizes constraints, evaluates them step-by-step, and produces an auditable validation log of constraint violations with associated evidence; an LLM-based judge uses this log to localize the critical step and category.
\tool improves step localization by 75\% on average over prior work, while providing failure category attribution.
\end{abstract}

\section{Introduction}
Large Language Model (LLM)-based agents are increasingly \emph{acting autonomously} without a human in the loop.
Modern agentic systems plan, coordinate, call tools, and execute complex real-world tasks.
These agents are no longer confined to sandboxes; they are being deployed in high-stakes environments such as recruiting~\citep{linkedin_hiring_assistant}, web browsing~\citep{fourney2024magentic-one}, and production cloud-service operations~\citep{zhang2024flash}. 
This ability to operate with minimal guidance dramatically amplifies human productivity ~\citep{NBERw31161}.
However, the same autonomy that enables speed also makes these agentic systems difficult to debug: trajectories are long-horizon, tool outputs are noisy, and failures can propagate through side effects before they are observed. 
Consequently, ensuring the robustness of AI agents is not an option, but a prerequisite for their safe adoption in the real-world.

\if
Recent work has begun to formalize \emph{failure attribution} for multi-agent systems, such as identifying the failure-responsible agent and the first error step in the log~\citep{zhang2025agent}.
This is an important step towards automation: knowing \emph{who} acted and \emph{when} the first mistake occurred provides useful debugging signal. 
It localizes a failure to a particular agent in the trace, but provides limited insight into \emph{what constraint was violated} and \emph{why the failure occurred}.
\fi
\begin{figure}[t]
  \centering
\includegraphics[width=\columnwidth]{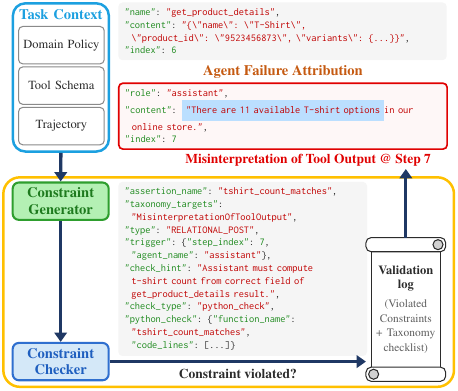}
  \vspace{-5mm}
  \caption{Given a failed trajectory, domain policy, and tool schema, \tool outputs the critical failure step, a root-cause category, and an auditable constraint violation.}
  \label{fig:agentrx}
\end{figure}
In this paper, our focus is on: \emph{root-causing and localizing the critical failure which prevented the Agent from successful task completion}.
We define \emph{critical failure} as the first unrecoverable failure by any agent in an execution trajectory. 
To understand and characterize such failures, we manually analyze failed agent executions and introduce a benchmark of 170 trajectories across 11 diverse task settings, including structured API workflows, incident management, and open-ended web/file tasks.
Each trajectory includes the full execution trace (messages, tool calls, tool outputs, and observable environment state) and is annotated with the \emph{critical step} and a \emph{failure category} capturing why the run ultimately fails.
Our grounded-theory-based annotation process also yields a cross-domain failure taxonomy for root-cause attribution.

\mypara{Challenges}
However, manual failure attribution is expensive for long-horizon trajectories and hard to scale for agents serving millions of daily users.
It requires not only inspecting long trajectories, but also domain expertise to understand the system prompt, available tools, domain policies, user guardrails, and expected state transitions.
This creates a need for automated validation of agent executions.
In software engineering, contracts make interface misuse and constraint violations explicit; agentic systems need an analogous notion of correctness, but one stated over executions: ambiguous intents, messages, tool calls, and state updates.

\mypara{\tool}
To operationalize this view, we design \tool, a \emph{domain-agnostic} automated framework for AI agents that
(1) normalizes heterogeneous multi-agent logs into a common trajectory representation,
(2) synthesizes \emph{constraints} from tool schemas, domain policies, and the observed trajectory, and
(3) produces a step-indexed validation log of violated obligations with supporting evidence, which a downstream judge uses to localize the critical failure step and assign a category.
%
%
As shown in Figure~\ref{fig:agentrx}, after the assistant agent calls \textsc{get\_product\_details} and reports ``11 available T-shirt options,'' \tool synthesizes a relational constraint \textsc{tshirt\_count\_matches} that recomputes the count from the tool response and checks agreement with the assistant’s interpretation.
%
The resulting violation includes the exact trajectory window that contradicts the assistant's claim, helping the judge localize the decisive failure.

\mypara{Evaluation}
We evaluate \tool on our proposed benchmark and prior failure-analysis datasets~\cite{zhang2025agent,cemri2026multi}.
\tool beats state of the art and our baselines on both localizing the first unresolved critical failure step and categorizing the root-cause. 
Our results clearly indicate that trajectory-level constraints are a useful signal for detecting failures. 
Our contributions include: (i) A manually annotated open-source dataset of \textbf{170 failed trajectories across diverse tasks}--including step localization and failure attribution ($\kappa$=0.89)\footnote{Benchmark: \url{https://huggingface.co/datasets/microsoft/AgentRx}.}.
(ii) a \textbf{novel domain-agnostic failure diagnosis framework} that combines constraint synthesis and LLM-based adjudication to localize agent failures.\footnote{Code: \url{https://github.com/microsoft/agentrx}} (iii) A \textbf{cross-domain failure taxonomy derived through grounded theory coding}, capturing 9 root-cause categories that generalize across diverse domains, and (iv) Extensive experiments showing 75\% average improvement in failure localization over prior work, and 30\% improvement in root-cause categorization over our own judge baselines.


\begin{figure*}[t]
  \centering
  \includegraphics[width=0.87\textwidth]{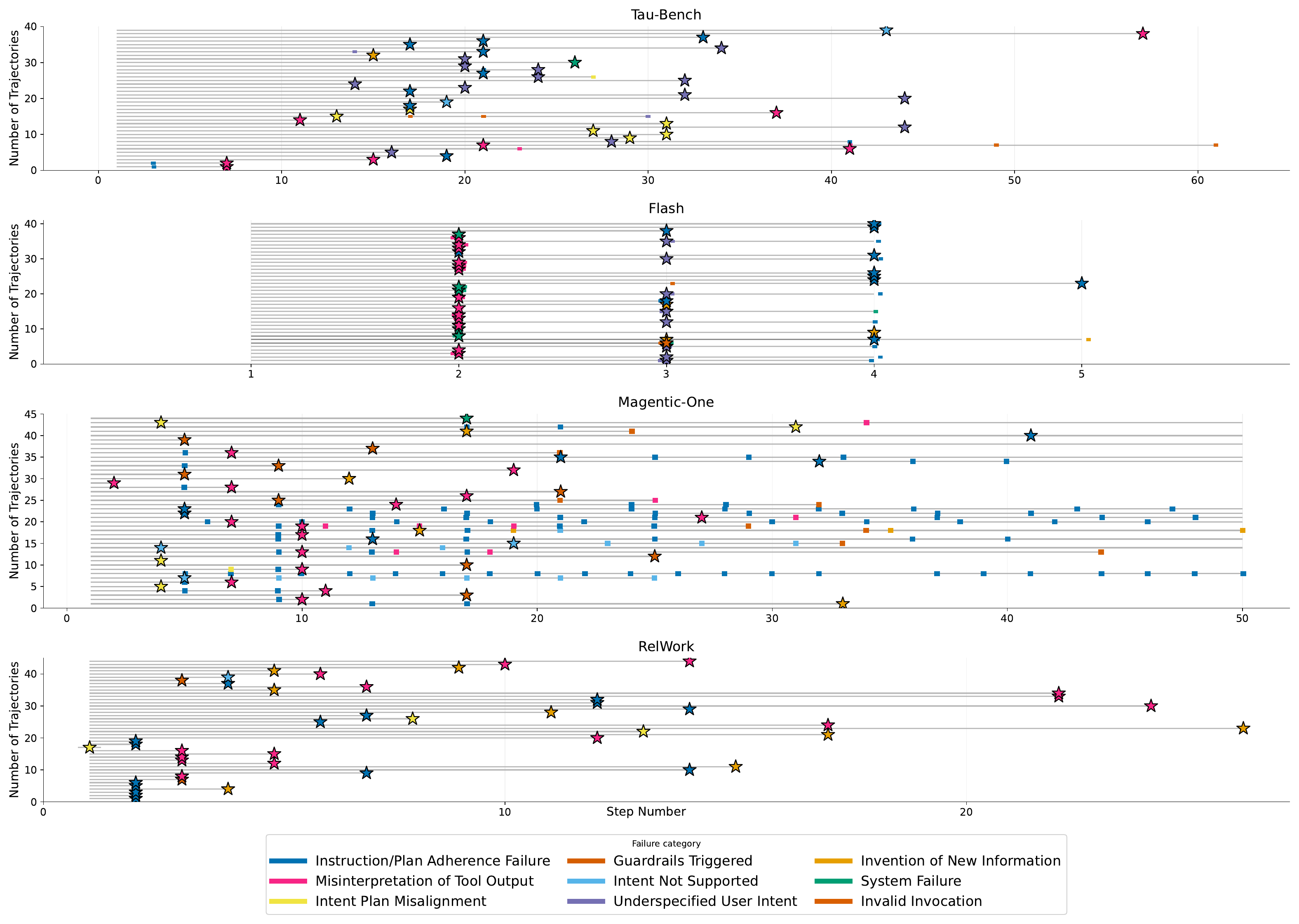}
  \caption{\textbf{Failure timelines per trajectory across domains.}
  Each horizontal line is one trajectory. Colored ticks mark detected failures at the corresponding step number (labeled with the category); the star indicates the root-cause step. Magentic-One has 295 total failures, with 68\% of trajectories containing at least two failures; Flash has 55 total failures, and $\tau$-bench has 52.}
  \label{fig:failure-timelines}
  \vspace{-2mm}
\end{figure*}
\section{The \tool Benchmark}
We are the first to introduce a benchmark for \emph{attribution of the first unrecoverable critical failure} in AI agents. 
Our benchmark contains 170 failed trajectories across 11 diverse task settings, spanning structured API workflows, production incident management, software-development agents, mathematical reasoning, and open-ended web/file tasks.
These settings vary along three axes: single-agent vs.\ multi-agent execution, structured APIs vs.\ open-ended environments, and recoverable vs.\ unrecoverable failure chains.
%
For each setting, we sample failed trajectories and use them to construct our failure-attribution benchmark.

\mypara{$\tau$-bench}
$\tau$-bench \citet{yao2024taubenchbenchmarktoolagentuserinteraction} is a tool-calling benchmark for retail tasks, with an LLM-simulated user and a single agent equipped with domain-specific APIs and policy guidelines.
In a typical $\tau$-bench task, the agent can cancel or modify pending orders, return or exchange delivered orders, update user addresses, and answer product queries.
We run 115 $\tau$-bench retail tasks using \textsc{gpt-4o} with one trial and analyze the 39 failed trajectories.

\mypara{Flash}
Flash~\citet{zhang2024flash} is a workflow automation agent for diagnosing recurring incidents in production cloud services. 
It is a multi-agent system in which specialized agents execute a troubleshooting guide for the incident being investigated.
The guide involves steps like querying for clusters in a region that might be affected and also mitigation steps based on whether the incident is a false alarm or not.
We analyze 42 Flash trajectories containing at least one agent failure.

\mypara{Magentic-One}
Magentic-One~\citet{fourney2024magentic-one} is a generalist multi-agent system for open-ended web and file-based tasks. 
It uses an orchestrator to coordinate specialized agents for web browsing, file navigation, and code execution.
We randomly sample 44 failed multi-agent trajectories from the Who\&When dataset~\citep{zhang2025agent}.
%


\mypara{RelWork}
We sample traces from MAST~\citep{cemri2026multi} and Who\&When~\citep{zhang2025agent}
by selecting five from each of their nine task domains.
Seven domains come from MAST and cover software-development agents, API/tool workflows, math reasoning, and open-ended web/file tasks.
The remaining two come from Who\&When: algorithm-generated and hand-crafted agent systems over GAIA/AssistantBench-style queries.
This yields 45 additional failed trajectories and brings the \tool benchmark to 170 trajectories in total.
This broadens the benchmark and lets us evaluate whether the taxonomy and attribution task generalize to agent-failure datasets that include long-horizon trajectories.
%
%
\subsection{Grounded Theory Annotation of Failures}
To systematically analyze the critical agent failures, we manually annotate trajectories using a grounded-theory open coding process~\citep{glaser2017discovery}, allowing failure categories to emerge from the traces rather than imposing a predefined label set.
%
%
%
%
Three annotators independently coded trajectories from the three domains.
Annotators read each trajectory step-by-step and wrote a structured \emph{coding memo} whenever a failure was observed.
We define a \emph{failure} as any point at which the agent did not make progress toward a correct outcome \emph{e.g.,} violating a policy guideline or executing an invalid tool call.
We follow a two-stage coding procedure:

\mypara{Phase 1: Exhaustive failure marking}
Annotators first marked \emph{all} failure steps, even if later steps dominate the final outcome. 
For each failure event, we recorded: (i) the step index; (ii) an \emph{open code}: a short, concrete description tightly grounded in the trace; and (iii) a reason for that failure. 
This stage captures early deviations and cascading failures, including ones the agent later recovers from.
%

\mypara{Phase 2: Critical failure identification}
We then identified the \emph{critical failure step}: the earliest failure in the causal chain from which the agent does not recover.
This step determines the root-cause category for the trajectory.
Throughout coding, annotators compared new codes against existing ones, reusing codes when possible and creating new definitions when needed.
We periodically merged related open codes into higher-level categories and refined their definitions.
We iterated until reaching \emph{theoretical saturation}, where new trajectories no longer introduced new failure phenomena.
The taxonomy is derived from three structurally different domains spanning both single-agent and multi-agent settings, each annotated independently by multiple annotators.
We then froze the category definitions and used closed coding to annotate remaining trajectories and re-code earlier ones; in particular, the RelWork trajectories were annotated entirely under the frozen taxonomy, so they serve as a held-out check that the categories transfer beyond the three domains that shaped them.

Each benchmark entry stores the trajectory identifier, the ordered list of failure events (step index, natural-language description, taxonomy label, rationale), and the critical failure; the event list gives the causal chain from the first unrecoverable failure to the terminal one.

%
%

\subsection{A Cross-Domain Failure Taxonomy}
\label{sec:taxonomy}
Prior work takes a system-level view of multi-agent failures, organizing failure modes by design, coordination, and verification stages~\citep{cemri2026multi}.
We develop a taxonomy that labels the critical failure:
\begin{enumerate}[leftmargin=*, label=\textbf{\arabic*.}, itemsep=2pt, topsep=2pt, parsep=0pt]
\item \textbf{Instruction/Plan Adherence Failure} The agent fails to follow required directions or the agreed plan by ignoring directives. This covers both under-execution and over-execution (unplanned actions) that violate the plan, domain policy, or orchestrator constraints.
\item \textbf{Invention of New Information} The agent introduces, removes, or alters information that is not grounded in any available input, context, or tool output.
This includes fabricating unsupported facts, hallucinated details, or omitting relevant information without justification. 
\item \textbf{Invalid Invocation} The agent issues an invalid tool call, e.g., malformed inputs, missing arguments, or values that fail schema validation.
\item \textbf{Misinterpretation of Tool Output} The agent incorrectly reasons about its own or another agent's tool output, leading to incorrect assumptions or actions.
\item \textbf{Intent–Plan Misalignment} The agent misinterprets the user's goal or constraints and produces an incorrect plan.
\item \textbf{Under-specified User Intent} The agent was unable to complete the task due to lack of complete information at that point in the trajectory.
\item \textbf{Intent Not Supported} The agent is asked to perform an action for which there is no tool.
\item \textbf{Guardrails Triggered} The agent's execution is blocked by safety policies or external access restrictions.
\item \textbf{System Failure} The agent faces a system connectivity issue while calling a particular tool like an API endpoint not being reachable.
\end{enumerate}

\mypara{Taxonomy Design}
%
%
%
The design principles for our failure taxonomy are: (i) \textbf{Mutual exclusivity at the critical step.} While multiple failure modes may appear across a trajectory, the critical failure step receives exactly one category label. This label captures the primary reason that the step results in an unrecoverable failure; (ii) \textbf{Disambiguation criteria.} Categories such as \emph{Instruction/Plan Adherence Failure}, where the agent deviates from the correct plan, and \emph{Intent--Plan Misalignment}, where the agent forms the wrong plan from the correct intent, are distinguished by whether the plan itself is correct, operationalized by the taxonomy checklist's yes/no questions (\autoref{app:checklist}); and 
(iii) \textbf{Extensibility.} 
The taxonomy is embedded as a fixed set in the prompt; extending it requires only: (a) defining new category questions in the checklist $\mathsf{K}$, and (b) adding them to the judge prompt.

\mypara{Inter-annotator Agreement}
We assessed inter-annotator agreement on 30 trajectories from $\tau$-bench, Flash, and Magentic-One, each independently annotated by three annotators using the same taxonomy and guidelines.
%
%
Because Cohen's $\kappa$ is defined for two annotators, we compute it pairwise across annotator pairs and report the mean.
The pairwise $\kappa$ scores were 0.93, 0.87, and 0.9, resulting in a mean $\kappa$ of 0.89. 
%
%
This corresponds to near perfect agreement (0.8-1.0), indicating that the taxonomy is well defined and used consistently. 
%
%
We manually reviewed disagreements to refine the taxonomy definitions; at the critical step, labels are mutually exclusive, with ambiguous cases resolved using a taxonomy checklist.

\mypara{Annotation Cost}
The manual effort required for these annotations is substantial, especially for long-horizon trajectories.
$\tau$-bench trajectories have median length 36 steps (range 20--62), Flash has median length 3 (range 2--6), Magentic-One has median length 33 (range 5--130), and RelWork has median length 10 (range 4--37).
Flash additionally averages 8 substeps per step.\footnote{Flash steps are defined using the Troubleshooting Guide for consistency; each step expands into multiple substeps.}
Annotators spent, on average, 20 minutes per $\tau$-bench trajectory, 22 minutes per Flash and RelWork trajectory, and 24 minutes per Magentic trajectory. %
Across 170 trajectories, this amounts to 62.5 total human hours and is challenging to scale. %
This motivates the need for automated failure attribution with \tool.

\begin{table}[]
 \caption{Distribution of \emph{critical failure} labels across domains. We normalize per domain (percent of failed trajectories)}
 \label{tab:tax-dist}
 \centering
 \small
 \setlength{\tabcolsep}{1pt}
 \resizebox{\columnwidth}{!}{%
 \begin{tabular}{lcccc}
 \toprule
 \textbf{Root-cause category} & \textbf{$\tau$-bench} & \textbf{Flash} & \textbf{Magentic} & \textbf{RelWork} \\
 \midrule
  Instr./Plan Adherence & 20.5 & 23.8 & 18.2 & 33.3 \\
 Invention of Information & 2.6 & 9.5 & 9.1 & 17.8 \\
 Invalid Invocation & 0 & 0 & 0 & 2.2 \\
 Misinterpretation of Output & 20.5 & 33.3 & 34.1 & 33.3 \\
 Intent-Plan Misalignment & 15.4 & 0 & 9.1 & 6.7 \\
 Under-specified Intent & 33.3 & 19 & 0 & 0 \\
 Intent Not Supported & 5.1 & 0 & 6.8 & 2.2 \\
 Guardrails Triggered & 0 & 0 & 20.5 & 2.2 \\
 System Failure & 2.6 & 14.3 & 2.3 & 0 \\
 \midrule
 \textbf{\# failed trajectories} & \textbf{39} & \textbf{42} & \textbf{44} & \textbf{45} \\
 \bottomrule
 \end{tabular}}
 \end{table}

\section{AgentRx Framework Formulation}
\label{sec:agentrx}

\mypara{Setup}
\tool takes as input (i) a toolset $\mathcal{T}$ with input--output schemas, (ii) an optional domain policy $\Pi$, and (iii) a failed trajectory $T=\langle s_1,\dots,s_n\rangle$.
%
%
Each step $s_k$ contains multiple logged events with fields such as agent name, tool name, step index, tool output or natural language conversational content.
%
%
%
Since log formats differ across domains, we first normalize each trajectory into a common intermediate representation (IR).
Let $T_{\le k}$ denote the trajectory prefix up to and including step $k$.
%
%
\tool generates \emph{executable constraints} from the tool schema, domain policy, and trajectory prefixes, then evaluates them step-by-step to produce an auditable validation log $\mathsf{V}$.
%
%
An LLM-based judge uses $\mathsf{V}$ and a taxonomy checklist $\mathsf{K}$ to predict the root-cause step $\hat{s}$ and category $\hat{y}$.

\mypara{Taxonomy checklist}
We manually compile our failure taxonomy into a semantic checklist $\mathsf{K}$: for each category $y$ in the taxonomy,
$\mathsf{K}(y)$ specifies a small set of targeted yes/no questions (and brief decision criteria) that operationalize
the category definition.
%

\subsection{Global and Dynamic Constraint Synthesis}
\label{sec:generation}
\tool synthesizes constraints in two stages:
\emph{Global constraints} $\mathcal{C}^G$ are synthesized once from the tool schema and, when available, the domain policy, and stored in a global store $\mathsf{G}$.
\emph{Dynamic constraints} $\mathcal{C}^D_k$ are synthesized at each step from the task instruction $I$ and the observed prefix $T_{\le k}$, using $\mathsf{G}$ as context, and stored in a local store $\mathsf{L}_k$.
Intuitively, global constraints encode domain rules, while dynamic constraints encode trajectory-specific rules and
prior observations (e.g., constraints that must hold given earlier tool outputs).
We define the constraints available at step $k$ as the union $\mathcal{C}_k := \mathcal{C}^G \cup \mathcal{C}^D_k$.
In practice, constraint synthesis emits constraints in a lightweight schema to standardize checking across domains (\autoref{sec:schema}).

\subsection{Guarded Constraints and Evaluation}
\label{sec:constraints}
A constraint $C$ generated by \tool comprises (i) a guard $G_C(T_{\le k},s_k)\in\{0,1\}$ that determines when the constraint applies and (ii) an assertion
$\Phi_C(T_{\le k},s_k)\in\{\textsc{sat},\textsc{viol}\}$ that returns a binary verdict $v$ when evaluated.
Guards capture structural conditions such as whether step $k$ calls a particular tool or invokes a particular agent.
Let $g_{C,k}=G_C(T_{\le k},s_k)$ and $\phi_{C,k}=\Phi_C(T_{\le k},s_k)$.
Evaluation at step $k$ is:
\[
\textsc{Eval}_C(k)=
\begin{cases}
(\textsc{skip},\emptyset) & \text{if } g_{C,k}=0,\\
(\phi_{C,k},e) & \text{otherwise.}
\end{cases}
\]
Constraints are checked either programmatically over structured fields (e.g., schema validity, equality, membership), or semantically as natural-language predicates evaluated using an LLM-based checker.
Both return a verdict--evidence pair $(v,e)$; examples appear in~\autoref{app:constraint-gen}.

%
%
 
%

\subsection{Validation Log}
\label{sec:verification}
At step $k$, \tool evaluates each constraint $C \in \mathcal{C}_k$ whose guard applies, i.e., $G_C(T_{\le k}, s_k)=1$.
We record each violated constraint and its supporting evidence in a step-indexed validation log:
\[
\begin{aligned}
\mathsf{V} \;:=\; \{(k,C,e)\;|\;& C\in \mathcal{C}_k,\ G_C(T_{\le k},s_k),\\
& (\textsc{viol},e)=\textsc{Eval}_C(k)\}.
\end{aligned}
\]
$\mathsf{V}$ is step-indexed, auditable, and directly links each violation to supporting evidence.

\subsection{Judge for Root Cause Attribution}
\label{sec:judge}
Given the task instruction $I$, trajectory $T$, checklist $\mathsf{K}$, and validation log $\mathsf{V}$, an LLM-based judge outputs a critical failure step $\hat{s}$ and category $\hat{y}$.
The judge selects the first step whose violation evidence is sufficient to explain why the run fails with respect to $I$.
Violations serve as diagnostic evidence rather than hard requirements: the judge may override them when the trajectory context indicates otherwise.
The judge sets $\hat{y}$ based on the taxonomy labels associated with the violation(s),
and emits a short rationale citing the specific reason for selecting the failure category.


%
%

%

%
\section{Experimental Evaluation}
\label{sec:experiments}
We evaluate \tool as a \emph{root-cause diagnosis} system for agent trajectories on our \tool benchmark.
We measure whether it can localize the first unrecoverable critical step and assign the correct root-cause category.
Our experiments test accuracy, cross-domain and cross-model generalization, contribution of validation signals, and robustness under relaxed metrics.
%

%
%

\subsection{Experimental Settings}
\mypara{Metrics}
We evaluate step localization using \emph{exact critical-step accuracy} (the fraction of trajectories where the predicted step exactly matches the annotated step), Acc@$\pm r$ for $r\in\{1,3,5\}$, and average step distance, which measures how far predictions are from the annotated step.
We evaluate category prediction using \emph{critical, any-failure, earliest, and terminal accuracy}, comparing the predicted category against the annotated critical failure, any annotated failure, the first observed failure, and the terminal failure, respectively.
The judge may also output \emph{inconclusive} when no grounded critical failure can be identified.










\mypara{Experimental settings}
We run each configuration $n{=}3$ times and report mean $\pm$ standard deviation to account for sampling noise and judge variability (Table~\ref{tab:ablations}).
%
%
Unless otherwise stated, \tool uses both programmatic constraints and semantic constraints.
Domain policies are optional: $\tau$-bench provides a policy document, while Flash and Magentic-One provide none, so their constraints derive from tool schemas and the observed trajectory alone.
We use \textsc{gpt-5} as the default model.
%
%
%
(Results using o3 and DeepSeek-V3.2 in \autoref{tab:o3-ablation} and \autoref{tab:deepseek}).
In Table~\ref{tab:ablations}, \emph{Baseline} denotes the judge without \tool-added evidence, while \emph{Baseline+Vio.} and related settings add validation signals incrementally.

\subsection{Comparison to Prior Benchmarks}
We compare against prior work: Who\&When is the closest benchmark for automated failure attribution, while MAST is the closest taxonomy-level analysis of multi-agent failures~(\autoref{app:related}).

\begin{table}[t]
\caption{
Comparison with related work. MAST~\cite{cemri2026multi} emphasizes broad multi-agent failure characterization, W\&W~\cite{zhang2025agent} focuses on responsible-agent and failure-step attribution, while AgentRx combines taxonomy-level analysis with evidence-grounded diagnosis of failed runs.
}
\label{app:related}
\centering
\footnotesize
\setlength{\tabcolsep}{4pt}
\renewcommand{\arraystretch}{1.10}

\newcommand{\cmark}{\checkmark}
\newcommand{\xmark}{--}
\newcommand{\pmark}{\(\sim\)}
\resizebox{\columnwidth}{!}{%
\begin{tabular}{@{}lccc@{}}
\toprule
\textbf{Criterion} & \textbf{MAST} & \textbf{W\&W} & \textbf{\tool} \\
\midrule
Failed-trace benchmark          & \cmark & \cmark & \cmark \\
Broad multi-agent coverage              & \cmark & \cmark & \cmark \\
Failure taxonomy                & \cmark & \xmark & \cmark \\
Cross-run failure analysis      & \cmark & \pmark & \cmark \\
Responsible-agent attribution   & \xmark & \cmark & \cmark \\
Failure-step localization       & \pmark & \cmark & \cmark \\
First unrecoverable failure     & \xmark & \pmark & \cmark \\
Evidence-grounded validation    & \xmark & \xmark & \cmark \\
\bottomrule
\end{tabular}}
\label{tab:related-work-comparison}
\end{table}

\mypara{Comparison to Who\&When}
Although W\&W annotates failed trajectories with responsible agents and failure steps, its target step is not always the critical failure under our definition.
In one trajectory~(\autoref{app:rootcause}), the Orchestrator recovers from the failure marked by W\&W: by switching from WebSurfer to FileSurfer, and only becomes unrecoverable after a later hallucinated file/path error (which our annotation catches).
%
%
This illustrates that our localization is a stricter debugging target.

Table~\ref{tab:failure-attribution-multidataset} compares \tool with W\&W across domains.
W\&W was designed for its own trajectory format and failure-attribution setting, so we used the \tool domain adapter to convert trajectories into the format required by W\&W.
We also modify the W\&W prompt to ask for the \emph{first unrecoverable} failure step, matching our task definition.
Across domains, \tool improves step localization over W\&W, with especially large gains on $\tau$-bench and RelWork.
\tool also improves failed-agent identification in most settings and, unlike W\&W, predicts a root-cause category.
\textbf{Our normalized IR and evidence-grounded validation improve attribution across heterogeneous agent trajectories.}

\begin{table}
\centering
\caption{\textbf{Failure attribution accuracy across datasets.}
We compare W\&W modified (W\&W$^{\ast}$) and \tool on failed-agent and critical-step accuracy. Mean $\pm$ std over 3 runs.}
\label{tab:failure-attribution-multidataset}
\small
\setlength{\tabcolsep}{2pt}
\renewcommand{\arraystretch}{0.8}

\begin{tabular}{llccc}
\toprule
\textbf{Dataset} & \textbf{Method} 
& \shortstack{\textbf{Agent}\\\textbf{Acc.}}
& \shortstack{\textbf{Step}\\\textbf{Acc.}}
& \shortstack{\textbf{Category}\\\textbf{Acc.}} \\
\midrule
 $\tau$-bench
 & W\&W$^{\ast}$  & $65 \pm 1.5$ & $18.8 \pm 5.3$ & -- \\
 & \tool & \textbf{66.7} & \textbf{42.7 $\pm$ 1.5} & \textbf{47 $\pm$ 1.5}\\
 \midrule
Magentic
& W\&W$^{\ast}$ & 61.4 $\pm$ 6.8 & 22.7 $\pm$ 3.9 & -- \\
& \tool          & \textbf{71.2 $\pm$ 1.1} & \textbf{36.4} & \textbf{46.2 $\pm$ 1.3} \\
\midrule
Flash 
& W\&W$^{\ast}$ & 16.1 $\pm$ 1.2 & 64.3 $\pm$ 3.9 & -- \\
& \tool & \textbf{82.5 $\pm$ 1.4} & \textbf{82.5 $\pm$ 1.4} & \textbf{65.9 $\pm$ 5} \\
\midrule
RelWork
& W\&W$^{\ast}$ & 57 $\pm$ 6.8 & 32.6 $\pm$ 1.3 & -- \\
& \tool          & \textbf{65.9 $\pm$ 5} & \textbf{61.5 $\pm$ 4.2} & \textbf{59.3 $\pm$ 2.1} \\

\bottomrule
\end{tabular}
\end{table}

\mypara{Comparison to MAST}
\begin{table*}[t]
\centering
\caption{\textbf{Effect of judge inputs and constraint generation.} We compare one-shot vs.\ step-by-step constraint generation, judge inputs (violations, taxonomy checklist, both) and judging protocols (Step-then-Category or All-at-Once).}
\label{tab:ablations}
\small
\setlength{\tabcolsep}{0.4pt}
\renewcommand{\arraystretch}{0.28}

\definecolor{blockgray}{gray}{0.88}

\begin{tabular*}{\textwidth}{@{\extracolsep{\fill}} l|c c|c c c c}
\toprule
& \multicolumn{2}{c|}{\textbf{Judge Baselines}} & \multicolumn{4}{c}{\textbf{\tool Ablations}} \\
\cmidrule(lr){2-3}\cmidrule(lr){4-7}
\textbf{Metric ($n{=}3$)}
& \textbf{Baseline} & \textbf{Step-then-Cat.}
& \textbf{Baseline+Vio.} & \textbf{Step-then-Cat.+Vio.} & \textbf{Taxonomy Checklist} & \textbf{Checklist+Vio.} \\
\midrule


\rowcolor{blockgray}
\textbf{Tau-Bench} & \multicolumn{6}{c}{\textbf{Step-by-Step Constraint Generation}} \\
Step-index acc. (\%) $\uparrow$  & 29.9 $\pm$ 1.2 & 31.6 $\pm$ 1.5 & \fbox{\textbf{42.7 $\pm$ 1.5}} & 38.5 $\pm$ 2.6 & 30.8 $\pm$ 2.1 & 39.3 $\pm$ 1.2  \\
Category acc. (\%) $\uparrow$    & 30.8 $\pm$ 2.1 & 32.5 $\pm$ 1.5 & 43.6 $\pm$ 1.2 & \fbox{\textbf{47 $\pm$ 1.5}} & 29.9 $\pm$ 2.4 & 47 $\pm$ 1.2 \\
Avg.\ step distance $\downarrow$ & 5.4  $\pm$ 0.1 & 5.5 $\pm$ 0.8 & \fbox{\textbf{3.8 $\pm$ 0.6}} & 4.2 $\pm$ 0.1 & 5.9  $\pm$ 0.3 & 3.9 $\pm$ 0.3 \\
\rowcolor{blockgray}
\textbf{Tau-Bench} & \multicolumn{6}{c}{\textbf{One-shot Constraint Generation}} \\
Step-index acc. (\%) $\uparrow$  & 29.9 $\pm$ 1.2 & 31.6 $\pm$ 1.5 & 41 $\pm$ 2.1 & 37.6 $\pm$ 1.2 & 30.8 $\pm$ 2.1 & \fbox{\textbf{42.7 $\pm$ 3.2}} \\
Category acc. (\%) $\uparrow$    & 30.8 $\pm$ 2.1 & 32.5 $\pm$ 1.5 & 41.9 $\pm$ 2.4 & 31.6 $\pm$ 1.2 & 29.9 $\pm$ 2.4 & \fbox{\textbf{43.6}} \\
Avg.\ step distance $\downarrow$ & 5.4  $\pm$ 0.1 & 5.5 $\pm$ 0.8  & 5.1 $\pm$ 0.4 & 5.3 $\pm$ 0.5 & 5.9 $\pm$ 0.3 & \fbox{\textbf{5 $\pm$ 0.3}} \\

\midrule

\rowcolor{blockgray}
 \textbf{Flash} & \multicolumn{6}{c}{\textbf{Step-by-Step Constraint Generation}} \\
Step-index acc. (\%) $\uparrow$  & 80.2 $\pm$ 1.4 & 76.2 $\pm$ 4.1 & 81.7 $\pm$ 1.4 & 75.4 $\pm$ 6.0 & \fbox{\textbf{82.5 $\pm$ 1.4}} & 80.2 $\pm$ 3.6 \\
  Category acc. (\%) $\uparrow$    & 58.7 $\pm$ 5.0 & 58.7 $\pm$ 1.4 & 59.5 $\pm$ 0.0 & \fbox{\textbf{65.9 $\pm$ 5.0}} & 62.7 $\pm$ 1.4 & 59.5 $\pm$ 2.4 \\
  Avg.\ step distance $\downarrow$ & 0.33 $\pm$ 0.04 & 0.39 $\pm$ 0.05 & 0.32 $\pm$ 0.04 & 0.42 $\pm$ 0.11 & \fbox{\textbf{0.29 $\pm$ 0.01}} & 0.34 $\pm$ 0.05 \\
\rowcolor{blockgray}
\textbf{Flash} & \multicolumn{6}{c}{\textbf{One-shot Constraint Generation}} \\
Step-index acc. (\%) $\uparrow$  & 80.2 $\pm$ 1.4 & 76.2 $\pm$ 4.1 & 81.0 $\pm$ 2.4 & 75.4 $\pm$ 3.6 & \fbox{\textbf{82.5 $\pm$ 1.4}} & 80.2 $\pm$ 1.4 \\
 Category acc. (\%) $\uparrow$    & 58.7 $\pm$ 5.0 & 58.7 $\pm$ 1.4 & 53.2 $\pm$ 1.4 & 56.3 $\pm$ 7.7 & \fbox{\textbf{62.7 $\pm$ 1.4}} & 57.9 $\pm$ 1.4 \\
 Avg.\ step distance $\downarrow$ & 0.33 $\pm$ 0.04 & 0.39 $\pm$ 0.05 & \fbox{\textbf{0.29 $\pm$ 0.04}} & 0.42 $\pm$ 0.06 & 0.29 $\pm$ 0.01 & 0.30 $\pm$ 0.03 \\

\midrule

\rowcolor{blockgray}
\textbf{Magentic} & \multicolumn{6}{c}{\textbf{Step-by-Step Constraint Generation}}\\
Step-index acc. (\%) $\uparrow$  & 32.6 $\pm$ 1.3  & 23.5 $\pm$ 1.3 & 34.1 $\pm$ 1.9 & 34.1 & 32.6 $\pm$ 1.3 & \fbox{\textbf{35.6 $\pm$ 1.3}} \\
Category acc. (\%) $\uparrow$    & 35.6 $\pm$ 1.3 & 38.6 & 41.7 $\pm$ 1.3 & \fbox{\textbf{46.2 $\pm$ 1.3}} & 37.1 $\pm$ 1.3 & 42.4 $\pm$ 1.3 \\
Avg.\ step distance $\downarrow$ & 22.5 $\pm$ 3 & 16.9 $\pm$ 0.7 & 14.9 $\pm$ 1.9 & 13.3 $\pm$ 0.8 & 18.9 $\pm$ 2.4 & \fbox{\textbf{12.4 $\pm$ 0.6}} \\
\rowcolor{blockgray}
\textbf{Magentic} & \multicolumn{6}{c}{\textbf{One-Shot Constraint Generation}} \\
Step-index acc. (\%) $\uparrow$  & 32.6 $\pm$ 1.3 & 23.5 $\pm$ 1.3 & 34.8 $\pm$ 1.1 & 28 $\pm$ 2.1 & 32.6 $\pm$ 1.3 & \fbox{\textbf{36.4}}\\
Category acc. (\%) $\uparrow$    & 35.6 $\pm$ 1.3 & 38.6 $\pm$ 1.3 & 38.6 $\pm$ 1.9  & 39.4 $\pm$ 1.1 & 37.1 $\pm$ 1.3 & \fbox{\textbf{40.9 $\pm$ 1.9}} \\
Avg.\ step distance $\downarrow$ & 22.5 $\pm$ 3 & 16.9 $\pm$ 0.7 & 15.8 $\pm$ 1.6 & 14.2 $\pm$ 0.7 & 18.9 $\pm$ 2.4 & \fbox{\textbf{14.1 $\pm$ 0.3}}\\

\midrule
\rowcolor{blockgray}
\textbf{RelWork} & \multicolumn{6}{c}
{\textbf{Step-by-Step Constraint Generation}} \\
Step-index acc. (\%) $\uparrow$  & 51.9 $\pm$ 4.6 & 52.6 $\pm$ 1.3 & \fbox{\textbf{61.5 $\pm$ 4.2}} & 54.5 $\pm$ 3.2 & 55.6 $\pm$ 4.8 & 61.5 $\pm$ 2.1 \\
Category acc. (\%) $\uparrow$    & 46.7 $\pm$ 3.1 & 48.1 $\pm$ 2.6 & 55.6 $\pm$ 1.8 & 49.2 $\pm$ 2.8 & 51.9 $\pm$ 5.5 & \fbox{\textbf{59.3 $\pm$ 2.1}} \\
Avg.\ step distance $\downarrow$ & 2 $\pm$ 0.2 & 1.9 $\pm$ 0.2 & \fbox{\textbf{1.2 $\pm$ 0.3}} & 1.7 $\pm$ 0.4 & 1.5 $\pm$ 0.1 & 1.8 $\pm$ 0.4 \\
\rowcolor{blockgray}
\textbf{RelWork} & \multicolumn{6}{c} 
{\textbf{One-shot Constraint Generation}} \\
 Step-index acc. (\%) $\uparrow$  & 51.9 $\pm$ 4.6 & 52.6 $\pm$ 1.3 & 47.4 $\pm$ 5.8 & 48.9 $\pm$ 3.6 & \fbox{\textbf{55.6 $\pm$ 4.8}} & 50.4 $\pm$ 2.1 \\
Category acc. (\%) $\uparrow$    & 46.7 $\pm$ 3.1 & 48.1 $\pm$ 2.6 & 43.7 $\pm$ 2.8 & 49.6 $\pm$ 4.2 & 51.9 $\pm$ 6.8 & \fbox{\textbf{54.1 $\pm$ 1.0}} \\
Avg.\ step distance $\downarrow$ & 2 $\pm$ 0.2 & 1.9 $\pm$ 0.2 & 1.8 $\pm$ 0.1 & 1.9 $\pm$ 0.1 & \fbox{\textbf{1.5 $\pm$ 0.1}} & 1.8 $\pm$ 0.03 \\
 \bottomrule
 \end{tabular*}
 \end{table*}
MAST studies aggregate multi-agent failure patterns, while \tool diagnoses individual failed trajectories by localizing the first unrecoverable failure and grounding it in trace evidence.
%
%
%
%
The taxonomies are not one-to-one: some \tool labels have only partial MAST analogs, while Invalid Invocation, Intent Not Supported, Guardrails Triggered, and System Failure do not appear as MAST labels.
Since MAST does not provide gold labels for our critical-step/root-cause task, we use it as a generalization stress test rather than an accuracy benchmark.
Running \tool without code changes on all 1,022 MAST trajectories yields a critical step, root-cause category, and supporting violation evidence for every trace; porting to a new agent system requires only a log-format adapter to the IR.
%
%
The resulting profiles vary by framework: Plan Adherence is dominant overall, while Invalid Invocation, Misinterpretation of Tool Output, and System Failure appear prominently in specific frameworks (\autoref{fig:mast-heatmap}); 7.6\% of traces receive \tool labels with no MAST counterpart.
 %
 Step localization also varied: AppWorld failed early, while Magentic-One failed later, consistent with MAST's observation that verification and termination failures are common (\autoref{fig:mast-violin}).
%


\subsection{Impact of Violations and Checklist}
Table~\ref{tab:ablations} compares \tool against a judge-only baseline that sees the trajectory and taxonomy prompt, but no \tool-generated validation evidence.
We isolate three added signals: violation evidence alone (Baseline+Vio.), checklist alone (Taxonomy Checklist), and checklist+violations (Checklist+Vio.).
%
%
%
Violation evidence gives the clearest gains on $\tau$-bench and RelWork.
The checklist is most useful on Flash, suggesting that \textbf{semantic structure can outweigh sparse or noisy violations evidence.}
On Magentic, combining checklist and violations gives the best step accuracy.
%
\textbf{Overall, violations and the taxonomy checklist are complementary: violations help localize grounded errors, while combining both signals gives reliable attribution across domains.}

\begin{table}[t]
\centering
\caption{\textbf{Per-category accuracy on $\tau$-bench} (fraction of each category's judgements labeled correctly), Taxonomy Checklist vs.\ Checklist+Vio.\ (step-by-step), pooled over $n{=}3$ runs.}
\label{tab:per-category}
\small
\begin{tabular}{@{}lccr@{}}
\toprule
\textbf{Category} & \textbf{Checklist} & \textbf{+Vio.} & $\Delta$ \\
\midrule
System Failure        & 0.0  & 66.7  & +66.7 \\
Intent--Plan Misalignment   & 5.6  & 61.1  & +55.5 \\
Misinterpretation of Output & 66.7 & 100.0 & +33.3 \\
Instruction/Plan Adherence     & 75.0 & 66.7  & $-$8.3 \\
Invention of Information    & 0.0  & 0.0   & 0 \\
Intent Not Supported  & 0.0  & 0.0   & 0 \\
Under-specified Intent   & 0.0  & 0.0   & 0 \\
\bottomrule
\end{tabular}
\end{table}

\mypara{Per-category attribution}
\autoref{tab:per-category} decomposes the $\tau$-bench Taxonomy Checklist and Checklist+Vio.\ results of \autoref{tab:ablations} by ground-truth category.
Violation evidence lifts three categories: System Failure and Intent--Plan Misalignment, which the checklist-only judge almost never predicts, and Misinterpretation of Tool Output.
Instruction/Plan Adherence drops slightly because the checklist-only judge uses it as a catch-all: with violation evidence, those failures are attributed to their actual causes instead.
\subsection{Impact of Constraint Generation}

We evaluate two constraint-generation strategies: \emph{one-shot}, which generates constraints once from the full trajectory, and \emph{step-by-step}, which generates constraints from each prefix $T_{\le k}$.
One-shot is more cost-efficient, while step-by-step better matches online diagnosis by conditioning only on information available up to step $k$.
As shown in \autoref{tab:ablations}, step-by-step constraint generation consistently outperforms one-shot generation across domains.
The advantage is most visible on longer trajectories: on Magentic, step-by-step Checklist+Vio.\ reduces average step distance from 22.5 to 12.4.
By contrast, Flash trajectories are shorter, and the two strategies perform more similarly.
%
%
%
%
%
%
%
%
This trend aligns with trajectory length statistics in \autoref{tab:token_stats}: longer traces are more vulnerable to context dilution under one-shot generation, \textbf{making localized step-by-step constraint generation more reliable.}
%
%
%
\begin{table}
 \centering
 \caption{Mean token and characters per trajectory.}
 \label{tab:token_stats}
 \small
 \setlength{\tabcolsep}{0.5pt}
 \renewcommand{\arraystretch}{0.4}
 \begin{tabular}{lcccc}
 \toprule
 \textbf{Metric} & $\boldsymbol{\tau}$-\textbf{Bench} & \textbf{Flash} & \textbf{Magentic} & \textbf{RelWork} \\
 \midrule
 \textbf{Avg tokens / step} & 133 & 169 & 330 & 393 \\
 \textbf{Avg chars / step}  & 480 & 930 & 1280 & 1711 \\
 \textbf{Avg tokens / trajectory}  & 4889 & 6415 & 16484 & 5837 \\
 \bottomrule
 \end{tabular}
 \end{table}
%
%
This tradeoff is also reflected in the cost.
Under the GPT-5 pricing assumptions used in our experiments, the judge-only baseline costs USD 0.05 per trajectory, while the one-shot \tool configuration costs USD 0.18 per trajectory, about 3.6$\times$ higher.
%
Most of this overhead comes from constraint generation, which accounts for roughly 81\% of the total cost; the final judge stage is comparable to the baseline.
Thus, \textbf{one-shot generation is a cost-efficient option when full-trajectory context is sufficient.}
Large-scale monitoring can use cheaper modes, while detailed debugging may justify step-by-step analysis.

\begin{table*}
\centering
\caption{\textbf{Step and category accuracy under tolerance. Mean $\pm$ std over $n{=}3$ runs.}}
\label{tab:step_cat_metrics_per_domain_transposed}
\small
\setlength{\tabcolsep}{1.5pt}
\renewcommand{\arraystretch}{0.7}

\begin{tabularx}{\textwidth}{@{}l *{5}{>{\centering\arraybackslash}X} | *{4}{>{\centering\arraybackslash}X}@{}}

\toprule
\textbf{Domain / Method}
& \textbf{\makecell{Step\\Acc. $\uparrow$}}
& \textbf{\makecell{Acc@\\$\pm1$ $\uparrow$}}
& \textbf{\makecell{Acc@\\$\pm3$ $\uparrow$}}
& \textbf{\makecell{Acc@\\$\pm5$ $\uparrow$}}
& \textbf{\makecell{Avg.\\Dist. $\downarrow$}}
& \textbf{\makecell{Critical\\Cat. $\uparrow$}}
& \textbf{\makecell{Any\\Cat. $\uparrow$}}
& \textbf{\makecell{Earliest\\Cat. $\uparrow$}}
& \textbf{\makecell{Terminal\\Cat. $\uparrow$}} \\
\midrule
 $\tau$-Bench (Baseline)   & 29.9 $\pm$ 1.2 & 41.0 $\pm$ 2.1 & 53.8 $\pm$ 2.1 & 68.4 $\pm$ 3.2 & 5.4 $\pm$ 0.14 & 30.8 $\pm$ 2.1 & 42.1 $\pm$ 5.7 & 36.8 $\pm$ 7.4 & 33.3 $\pm$ 3.3 \\
 
$\tau$-Bench (\tool)   & 42.7 $\pm$ 1.2 & 48.7 $\pm$ 4.2 & 61.5 $\pm$ 3.6 & 68.4 $\pm$ 2.4 & 3.8 $\pm$ 0.6 & 47.0 $\pm$ 1.2 & 50.9 $\pm$ 3.3 & 43.0 $\pm$ 3.3 & 43.9 $\pm$ 3.3 \\
\addlinespace[1pt]
\midrule

Flash (Baseline)          & 80.2 $\pm$ 1.4 & 94.4 $\pm$ 1.1 & 100 & 100 & 0.2 & 58.7 $\pm$ 5 & 62.7 $\pm$ 4.4 & 53.2 $\pm$ 2.9 & 62.7 $\pm$ 4.4 \\
Flash (\tool)     &  82.5 $\pm$ 1.4 &  93.7 $\pm$ 1.4 & 97.6 & 100 & 0.2 & 65.9 $\pm$ 5.0 & 65.9 $\pm$ 5.0 & 65.9 $\pm$ 5.0 & 65.9 $\pm$ 5.0 \\
\addlinespace[1pt]
\midrule



Magentic (Baseline)  & 32.6 $\pm$ 1.3 & 43.2 $\pm$ 3.2 & 43.9 $\pm$ 4.3 & 52.3 $\pm$ 4.9 & 22.5 $\pm$ 3 & 35.6 $\pm$ 1.3 & 50.8 $\pm$ 2.1 & 39.4 $\pm$ 2.1 & 34.1 \\
Magentic (\tool)    & 36.4 & 47 $\pm$ 5.4 & 54.6 $\pm$ 4.9 & 59.8 $\pm$ 2.8 & 12.4 $\pm$ 0.6 & 46.2 $\pm$ 1.3  & 59.1 $\pm$ 2.1 & 43.2 $\pm$ 2.8 & 39.4 $\pm$ 1.1 \\
\midrule
 RelWork (Baseline)  & 51.9 $\pm$ 4.6 & 68.1 $\pm$ 3.8 & 84.4 $\pm$ 0.0 & 88.9 $\pm$ 1.8 & 2.0 $\pm$ 0.2 & 46.7 $\pm$ 3.1 & 46.7 $\pm$ 3.1 & 46.7 $\pm$ 3.1 & 46.7 $\pm$ 3.1 \\
 RelWork (\tool)      & 61.5 $\pm$ 4.2 & 76.3 $\pm$ 2.1 & 92.6 $\pm$ 2.8 & 92.6 $\pm$ 2.8 & 1.2 $\pm$ 0.3 & 59.3 $\pm$ 2.1 & 59.3 $\pm$ 2.1 & 59.3 $\pm$ 2.1 & 59.3 $\pm$ 2.1 \\\bottomrule
\end{tabularx}
\end{table*}
%

%
%

%

\subsection{LLM-as-a-Judge Evaluation}
We compare two LLM-as-judge protocols: \emph{All-at-Once}, which predicts the critical step and category in one call, and \emph{Step-then-Category}, which first predicts the step and then categorizes that step.
%
%
All-at-Once is the default \tool setting.
In \autoref{tab:ablations}, Step-then-Category often improves category accuracy, including on $\tau$-bench, Flash, and Magentic, but it can compromise step localization because the category decision is conditioned on a potentially noisy first-stage step.


%
%
%

\subsection{Robustness Across Relaxed Metrics}

\textbf{Step accuracy increases as we allow more tolerance in the predicted step} (Acc@$\pm1$ $<$ Acc@$\pm3$ $<$ Acc@$\pm5$) as shown in \autoref{tab:step_cat_metrics_per_domain_transposed}.
For category prediction, we report: \emph{critical} accuracy; \emph{any-failure} accuracy, where predicted accuracy matches emph{any} category among the trajectory’s failure steps; \emph{earliest} accuracy, where it matches the first failure; and \emph{terminal} accuracy, where it matches the last failure.
Critical accuracy measures root-cause attribution, while any-failure accuracy captures the weaker but practical ability to flag a failure.
%
%
%
\textbf{\tool remains useful under both strict and relaxed failure-attribution metrics.}

\subsection{Qualitative Failure Analysis}
\label{sec:error-analysis}

We manually inspect cases where \tool produces an incorrect or misleading diagnosis. 
\paragraph{Surfaced violations can misdirect the judge.}
The violation evidence presented to the judge sometimes points toward the wrong type of failure. 
In a Flash example, the correct label is \emph{System Failure}, since the underlying problem is infrastructural: the Kusto query failed because of an endpoint issue. 
%
However, the violation log is dominated by invocation-oriented checks about predefined queries and correct cluster selection. 
To mitigate this, we have an optional reflection and pruning stage that asks the LLM to remove likely false-positive generated constraints before judging.

\paragraph{Surfaced violations can capture downstream symptoms.}
Surfaced violations may be real but downstream of the root cause, shifting the diagnosis to a secondary failure.
In Magentic-One task \texttt{1f975693-876d-457b-a649-393859e79bf3}, the correct label is \emph{Invention of New Information}: the Orchestrator hallucinated that a PDF was successfully downloaded and planned based on that false assumption. AgentRx's checklist focused instead on a late-trajectory guardrail block at step 51, where a \texttt{ResponsibleAIPolicyViolation} occurred and the Orchestrator emitted a \texttt{FINAL ANSWER} that was ungrounded. This shifted the diagnosis to \emph{Guardrails Triggered}, because the violation pointed at a downstream symptom.
%
%
%
%

\section{Related Work}
\label{sec:related}
\mypara{Benchmarks for Agent Evaluation}
Recent work has introduced benchmarks that cover tool execution, web interaction, and assistant behavior~\citep{barres2025tau,yao2022webshop,drouin2024workarena}.
AgentBench~\citet{liu2024agentbench} evaluates LLM agents across eight interactive environments that span code-centric, game-based, and web-based settings, including domains like operating systems and databases.
WebArena~\citet{zhou2024webarena} evaluates long-horizon web agents on real websites, testing whether they can convert language instructions into executable browser interactions.
GAIA~\citet{mialon2023gaia} targets general assistants with problems that are easy for humans but often require multi-step tool use, and has been used to evaluate generalist agent systems.
However, these benchmarks primarily report terminal success and do not include gold failure-attribution annotations.
Our dataset annotates the first unrecoverable critical step and assigns a failure category from a cross-domain taxonomy.
As discussed above, Who\&When is our closest comparison.

\mypara{Agent Reliability} 
\citet{koohestani2025agentguardruntimeverificationai} provides runtime assurance for AI agents by monitoring executions against learned behavioral models.
Recent work also translates plans into formal representations and applies model checking to validate alignment between an agent's behavior and its plan~\citep{ramani2025bridgingllmplanningagents,pmlr-v267-zhang25ds}.
These methods primarily target runtime enforcement or plan-conformance checks; our focus is to diagnose trajectories by extracting constraints from tools, policies, and the observed prefix, and use the resulting violations as evidence for failure attribution.
A separate line of work improves agent performance through feedback and self-correction.
Self-Refine~\citep{madaan2023selfrefineiterativerefinementselffeedback} formalizes an iterative loop in which an LLM generates critiques of its own outputs and applies revisions.
Critic~\citep{gou2024criticlargelanguagemodels} uses external tools like search and code execution as feedback to verify outputs.
%
%


\mypara{LLM Evaluation of Agent Trajectories}
%
~\citet{gu2025surveyllmasajudge} systematizes judge design and mitigation strategies for inconsistency and bias. 
Agent-as-a-Judge~\citet{zhuge2025agent} uses tool-using agents to evaluate other agents, arguing benefits over single-shot judges.
\tool is compatible with these paradigms, but shifts the judge from scoring to diagnosis by providing evidence.
\section{Conclusion}
Agents should not only be evaluated by whether they fail, but by where recovery becomes impossible.
We present a domain-agnostic framework that diagnoses failed trajectories by generating constraints and assigning root-cause categories.
%
%
\tool moves agent evaluation toward evidence-grounded debugging: explaining not just that an agent failed, but when and why.

\section{Limitations} 
Our failure taxonomy covers diverse domains and annotations from three independent annotators.
However, it may not cover all failure modes in other agentic domains and may require extension.
Sometimes \tool can be misled when validation signals are weak or contain false positives.
%
%
showing that noisy or downstream violation signals can misdirect the judge.
This motivates future work on identifying the smallest set of high-quality signals needed to separate true failures from noisy flags.
Although \tool relies on LLM calls, it reduces manual effort for failure attribution and is easier to scale.
\tool also offers cost-efficient one-shot constraint generation as an alternative.
We view both the framework and the dataset as a step toward systematic, evidence-based failure attribution in AI agents.

\section*{Ethical Considerations}
This paper is focused on making AI agents easier to debug. As agents increasingly run multi-step workflows: calling tools, changing state, and coordinating sub-agents failures are difficult. They can be wrong tool calls, skipped confirmations, misread outputs, or decisions that push the run onto a path it never recovers from. Our goal is to help developers pinpoint where the run first became unrecoverable and why, using evidence already present in execution traces.

The main upside is practical: faster debugging, clearer accountability, and safer iteration. A system that can surface the root cause and connect it to trace evidence can make it easier to measure reliability. The benchmark and taxonomy also encourage better instrumentation and more careful thinking about what should be logged and what should be checked.

There are also risks. Better diagnosis does not automatically mean agents are safe to deploy, and explanations can be over-trusted if people treat them as guarantees. Trajectory data can also contain sensitive information depending on the application, so anyone using these methods should handle logs carefully and follow appropriate privacy practices. 
Finally, a taxonomy learned from existing benchmarks can inherit their blind spots, so it should be treated as a starting point rather than a complete map of real-world failures.
Overall, we expect this work to be most useful as an Agentic debugging framework: it helps find problems earlier and fix them  proactively, but it does not replace careful testing.


\bibliography{example_paper}

@misc{koohestani2025agentguardruntimeverificationai,
      title={AgentGuard: Runtime Verification of AI Agents}, 
      author={Roham Koohestani},
      year={2025},
      eprint={2509.23864},
      archivePrefix={arXiv},
      primaryClass={cs.AI},
      url={https://arxiv.org/abs/2509.23864}, 
}

@misc{ramani2025bridgingllmplanningagents,
      title={Bridging LLM Planning Agents and Formal Methods: A Case Study in Plan Verification}, 
      author={Keshav Ramani and Vali Tawosi and Salwa Alamir and Daniel Borrajo},
      year={2025},
      eprint={2510.03469},
      archivePrefix={arXiv},
      primaryClass={cs.AI},
      url={https://arxiv.org/abs/2510.03469}, 
}

@misc{yao2024taubenchbenchmarktoolagentuserinteraction,
      title={$\tau$-bench: A Benchmark for Tool-Agent-User Interaction in Real-World Domains}, 
      author={Shunyu Yao and Noah Shinn and Pedram Razavi and Karthik Narasimhan},
      year={2024},
      eprint={2406.12045},
      archivePrefix={arXiv},
      primaryClass={cs.AI},
      url={https://arxiv.org/abs/2406.12045}, 
}

@inproceedings{zhang2025agent,
  title={Which Agent Causes Task Failures and When? On Automated Failure Attribution of LLM Multi-Agent Systems},
  author={Zhang, Shaokun and Yin, Ming and Zhang, Jieyu and Liu, Jiale and Han, Zhiguang and Zhang, Jingyang and Li, Beibin and Wang, Chi and Wang, Huazheng and Chen, Yiran and others},
  booktitle={International Conference on Machine Learning},
  pages={76583--76599},
  year={2025},
  organization={PMLR}
}

@book{glaser2017discovery,
  title={Discovery of grounded theory: Strategies for qualitative research},
  author={Glaser, Barney and Strauss, Anselm},
  year={2017},
  publisher={Routledge}
}

@article{cemri2026multi,
  title={Why do multi-agent llm systems fail?},
  author={Cemri, Mert and Pan, Melissa Z and Yang, Shuyi and Agrawal, Lakshya A and Chopra, Bhavya and Tiwari, Rishabh and Keutzer, Kurt and Parameswaran, Aditya and Klein, Dan and Ramchandran, Kannan and others},
  journal={Advances in Neural Information Processing Systems},
  volume={38},
  year={2026}
}

@inproceedings{zhuge2025agent,
  title={Agent-as-a-Judge: Evaluate Agents with Agents},
  author={Zhuge, Mingchen and Zhao, Changsheng and Ashley, Dylan R and Wang, Wenyi and Khizbullin, Dmitrii and Xiong, Yunyang and Liu, Zechun and Chang, Ernie and Krishnamoorthi, Raghuraman and Tian, Yuandong and others},
  booktitle={International Conference on Machine Learning},
  pages={80569--80611},
  year={2025},
  organization={PMLR}
}

@misc{gu2025surveyllmasajudge,
      title={A Survey on LLM-as-a-Judge}, 
      author={Jiawei Gu and Xuhui Jiang and Zhichao Shi and Hexiang Tan and Xuehao Zhai and Chengjin Xu and Wei Li and Yinghan Shen and Shengjie Ma and Honghao Liu and Saizhuo Wang and Kun Zhang and Yuanzhuo Wang and Wen Gao and Lionel Ni and Jian Guo},
      year={2025},
      eprint={2411.15594},
      archivePrefix={arXiv},
      primaryClass={cs.CL},
      url={https://arxiv.org/abs/2411.15594}, 
}

@misc{madaan2023selfrefineiterativerefinementselffeedback,
      title={Self-Refine: Iterative Refinement with Self-Feedback}, 
      author={Aman Madaan and Niket Tandon and Prakhar Gupta and Skyler Hallinan and Luyu Gao and Sarah Wiegreffe and Uri Alon and Nouha Dziri and Shrimai Prabhumoye and Yiming Yang and Shashank Gupta and Bodhisattwa Prasad Majumder and Katherine Hermann and Sean Welleck and Amir Yazdanbakhsh and Peter Clark},
      year={2023},
      eprint={2303.17651},
      archivePrefix={arXiv},
      primaryClass={cs.CL},
      url={https://arxiv.org/abs/2303.17651}, 
}

@techreport{fourney2024magentic-one,
author = {Fourney, Adam and Bansal, Gagan and Mozannar, Hussein and Tan, Cheng and Salinas, Eduardo and Zhu, Erkang (Eric) and Niedtner, Friederike and Proebsting, Grace and Bassman, Griffin and Gerrits, Jack and Alber, Jacob and Chang, Peter and Loynd, Ricky and West, Robert and Dibia, Victor and Awadallah, Ahmed and Kamar, Ece and Hosn, Rafah and Amershi, Saleema},
title = {Magentic-One: A Generalist Multi-Agent System for Solving Complex Tasks},
institution = {Microsoft},
year = {2024},
month = {November},
url = {https://www.microsoft.com/en-us/research/publication/magentic-one-a-generalist-multi-agent-system-for-solving-complex-tasks/},
number = {MSR-TR-2024-47},
}

@misc{gou2024criticlargelanguagemodels,
      title={CRITIC: Large Language Models Can Self-Correct with Tool-Interactive Critiquing}, 
      author={Zhibin Gou and Zhihong Shao and Yeyun Gong and Yelong Shen and Yujiu Yang and Nan Duan and Weizhu Chen},
      year={2024},
      eprint={2305.11738},
      archivePrefix={arXiv},
      primaryClass={cs.CL},
      url={https://arxiv.org/abs/2305.11738}, 
}

@article{barres2025tau,
  title={Evaluating Conversational Agents in a Dual-Control Environment},
  author={Barres, Victor and Dong, Honghua and Ray, Soham and Si, Xujie and Narasimhan, Karthik},
  journal={arXiv preprint arXiv:2506.07982},
  year={2025}
}

@inproceedings{mialon2023gaia,
  title={Gaia: a benchmark for general ai assistants},
  author={Mialon, Gr{\'e}goire and Fourrier, Cl{\'e}mentine and Wolf, Thomas and LeCun, Yann and Scialom, Thomas},
  booktitle={The Twelfth International Conference on Learning Representations},
  year={2023}
}

@inproceedings{zhou2024webarena,
  title={Webarena: A realistic web environment for building autonomous agents},
  author={Zhou, Shuyan and Xu, Frank F and Zhu, Hao and Zhou, Xuhui and Lo, Robert and Sridhar, Abishek and Cheng, Xianyi and Ou, Tianyue and Bisk, Yonatan and Fried, Daniel and others},
  booktitle={International Conference on Learning Representations},
  volume={2024},
  pages={15585--15606},
  year={2024}
}

@inproceedings{liu2024agentbench,
  title={Agentbench: Evaluating llms as agents},
  author={Liu, Xiao and Yu, Hao and Zhang, Hanchen and Xu, Yifan and Lei, Xuanyu and Lai, Hanyu and Gu, Yu and Ding, Hangliang and Men, Kaiwen and Yang, Kejuan and others},
  booktitle={International Conference on Learning Representations},
  volume={2024},
  pages={52989--53046},
  year={2024}
}

@article{yao2022webshop,
  title={Webshop: Towards scalable real-world web interaction with grounded language agents},
  author={Yao, Shunyu and Chen, Howard and Yang, John and Narasimhan, Karthik},
  journal={Advances in Neural Information Processing Systems},
  volume={35},
  pages={20744--20757},
  year={2022}
}

@inproceedings{drouin2024workarena,
  title={WorkArena: How Capable are Web Agents at Solving Common Knowledge Work Tasks?},
  author={Drouin, Alexandre and Gasse, Maxime and Caccia, Massimo and Laradji, Issam H and Del Verme, Manuel and Marty, Tom and Vazquez, David and Chapados, Nicolas and Lacoste, Alexandre},
  booktitle={International Conference on Machine Learning},
  pages={11642--11662},
  year={2024},
  organization={PMLR}
}

@techreport{NBERw31161,
 title = "Generative AI at Work",
 author = "Brynjolfsson, Erik and Li, Danielle and Raymond, Lindsey R",
 institution = "National Bureau of Economic Research",
 type = "Working Paper",
 series = "Working Paper Series",
 number = "31161",
 year = "2023",
 month = "April",
 doi = {10.3386/w31161},
 URL = "http://www.nber.org/papers/w31161",
}

@unpublished{zhang2024flash,
author = {Zhang, Xuchao and Mittal, Tanish and Bansal, Chetan and Wang, Rujia and Ma, Minghua and Ren, Zhixin and Huang, Hao and Rajmohan, Saravan},
title = {FLASH: A Workflow Automation Agent for Diagnosing Recurring Incidents},
year = {2024},
month = {October},
url = {https://www.microsoft.com/en-us/research/publication/flash-a-workflow-automation-agent-for-diagnosing-recurring-incidents/},
}

@InProceedings{pmlr-v267-zhang25ds,
  title = 	 {Position: Trustworthy {AI} Agents Require the Integration of Large Language Models and Formal Methods},
  author =       {Zhang, Yedi and Cai, Yufan and Zuo, Xinyue and Luan, Xiaokun and Wang, Kailong and Hou, Zhe and Zhang, Yifan and Wei, Zhiyuan and Sun, Meng and Sun, Jun and Sun, Jing and Dong, Jin Song},
  booktitle = 	 {Proceedings of the 42nd International Conference on Machine Learning},
  pages = 	 {82441--82459},
  year = 	 {2025},
  editor = 	 {Singh, Aarti and Fazel, Maryam and Hsu, Daniel and Lacoste-Julien, Simon and Berkenkamp, Felix and Maharaj, Tegan and Wagstaff, Kiri and Zhu, Jerry},
  volume = 	 {267},
  series = 	 {Proceedings of Machine Learning Research},
  month = 	 {13--19 Jul},
  publisher =    {PMLR},
  url = 	 {https://proceedings.mlr.press/v267/zhang25ds.html}
}

@misc{linkedin_hiring_assistant,
  author       = {{LinkedIn}},
  title        = {Hiring Assistant for {LinkedIn} Recruiter \& Jobs},
  howpublished = {\url{https://business.linkedin.com/talent-solutions/hiring-assistant}},
  year         = {2024},
  note         = {Accessed: 2025-01-28}
}

\appendix

\section{Taxonomy Checklist}
\label{app:checklist}
The taxonomy checklist $\mathsf{K}$ (\autoref{sec:agentrx}) is compiled by hand from the taxonomy. For each category it gives a decision criterion and a small set of yes/no questions; the judge is instructed to answer them for the candidate step before committing to a label. We reproduce it verbatim.

\newcommand{\chk}[1]{\vspace{-2pt}\begin{itemize}[leftmargin=*,itemsep=0pt,topsep=1pt,parsep=0pt]#1\end{itemize}}

\mypara{1. Instruction/Plan Adherence Failure}
Goal is correct, but the agent deviates from the required plan by ignoring directives and skipping steps despite having enough information. This covers both under-execution (missed steps) and over-execution (unplanned or unnecessary actions, e.g., extra tool calls) that deviate from the static plan, domain policy or orchestrator plan.
\chk{
\item Can you state the user's goal, and do the agent's intent and end goal match that goal (i.e., the agent is not solving the wrong problem)?
\item Was all the required information already available at this step (user intent, required context, prior tool outputs)?
\item Is there a step where the ground-truth/policy requires an action (tool call, question, confirmation, ordering) and the agent did something different (skipped it / reordered it / added extra unneeded action)?
}

\mypara{2. Invention of New Information}
The agent introduces, removes, or alters information that is not grounded in any available input, context, or tool output. This includes fabricating unsupported facts, hallucinating details, or omitting relevant information.
\chk{
\item Can you pinpoint the exact invented/altered/omitted claim, value, or assumption the agent used?
\item Is that claim absent from all evidence available up to that step (user text, provided context, tool outputs)?
\item Did the agent rely on that claim to decide an action or produce the failing conclusion (not just harmless wording)?
}

\mypara{3. Invalid Invocation}
Tool call fails because the request is ill-formed (missing args, wrong fields/types, malformed query, schema mismatch).
\chk{
\item At the failure step, did the agent attempt a tool call with a concrete invocation payload/arguments?
\item Does the tool/runtime explicitly report a parse/validation/schema/syntax error for that call (e.g., missing field, invalid type, cannot parse, malformed query)?
\item Is the error NOT primarily a network/timeout/service-unavailable/endpoint-unreachable issue (infra/connectivity)?
\item Is the error NOT primarily a CAPTCHA/login/paywall refusal (access/guardrail block)?
}

\mypara{4. Misinterpretation of Tool Output / Handoff Failure}
The agent incorrectly reasons about its own or another agent's tool output, leading to incorrect assumptions or actions. This also includes cases where the agent considered only partial tool output.
\chk{
\item Before (or at) the failure step, did the agent receive tool output or handoff output that is relevant to the failing decision?
\item Did the agent state or imply a specific reasoning derived from that tool output?
\item Does that reasoning contradict the tool output, omit a crucial part, or reflect a clear computation/logic error relative to the output?
}

\mypara{5. Intent--Plan Misalignment}
Agent misunderstands the user's intent/constraints and pursues the wrong objective or violates key constraints due to misunderstanding.
\chk{
\item Do the agent's actions/plan optimize for a different goal OR violate a key constraint (not a minor wording/format issue)?
\item Is the misalignment due to misunderstanding of intent/constraints (rather than missing required info from the user/context/tool outputs)?
\item Is the misalignment not primarily caused by a tool error (invalid invocation, infra failure, or access/guardrail block)?
}

\mypara{6. Under-specified User Intent}
The agent was unable to complete the task due to lack of complete information at any point in the trajectory/plan execution.
\chk{
\item Can you identify a specific missing piece of information that is required to proceed correctly (e.g., date, address, account id, item variant)?
\item Is that information absent from all evidence available up to that step (user text, provided context, and tool outputs)?
\item Did the agent fail because it proceeded without obtaining this information OR because it did not ask for it when needed?
}

\mypara{7. Intent Not Supported}
Requested action cannot be performed with available tools/capabilities.
\chk{
\item Is the user requesting an action that requires an external capability/tool (e.g., listen to audio, access a private system, perform a human action)?
\item Given the tool set available in this environment, is there no tool that can perform the requested action?
\item Is the failure not primarily caused by infrastructure/connectivity issues?
}

\mypara{8. Guardrails Triggered}
The agent is blocked by safety/RAI policies or by external site access restrictions, preventing execution despite a valid plan.
\chk{
\item Is there an explicit refusal/block signal (policy refusal, CAPTCHA, login required, 403, paywall, robots.txt, automation forbidden)?
\item Would the plan be feasible and correct if this block were removed (i.e., the agent is not pursuing the wrong goal/constraints)?
\item Is the failure not primarily due to malformed tool invocation (schema/syntax/args validation error)?
\item Is the failure not primarily due to infrastructure/connectivity issues (timeouts, endpoint unreachable)?
}

\mypara{9. System Failure}
The agent faces a system connectivity issue while calling a particular tool like an endpoint not being reachable.
\chk{
\item At the failure step, did the agent attempt a tool call or rely on a tool that should have been callable?
\item Is there an explicit infra/connectivity error signal (timeout, connection refused, DNS failure, endpoint unreachable, service unavailable, premature termination)?
\item Is the failure not primarily a parse/validation/schema/syntax error caused by malformed arguments?
}

\mypara{10. Inconclusive (use sparingly)}
None of 1--9 clearly apply; the judge must provide a custom category label.
\chk{
\item If labeling as 10, did you provide a non-empty \texttt{custom\_category} describing the failure type?
}
%


\begin{table*}[t]
  \centering
  \caption{\textbf{o3 judge ablation} (mean $\pm$ std, $n{=}3$; configurations as in \autoref{tab:ablations}). Violation evidence helps o3 where coverage is dense ($\tau$-bench, Flash); on the context-heavy domains (Magentic's long traces, RelWork's heavyweight steps, \autoref{tab:token_stats}) o3 is competitive without it, mirroring \autoref{tab:deepseek}.}
  \label{tab:o3-ablation}
  \resizebox{\textwidth}{!}{%
  \begin{tabular}{ll cccccc}
  \toprule
   & & \multicolumn{2}{c}{\textbf{Judge Baselines}} & \multicolumn{4}{c}{\textbf{AGENTRX Ablations}} \\
  \cmidrule(lr){3-4} \cmidrule(lr){5-8}
  Domain & Metric (n=3) & Baseline & Step-then-Cat. & Baseline+Vio. & Step-then-Cat.+Vio. & Taxonomy Checklist & Checklist+Vio. \\
  \midrule
  \multirow{3}{*}{$\tau$-bench}
    & Step-index acc. (\%) $\uparrow$  & 29.1 $\pm$ 1.2 & 28.2 $\pm$ 2.1 & 27.4 $\pm$ 1.2 & 21.4 $\pm$ 1.2 & 29.9 $\pm$ 2.4 & \fbox{\textbf{34.2 $\pm$ 3.2}} \\
    & Category acc. (\%) $\uparrow$    & 32.5 $\pm$ 1.2 & 31.6 $\pm$ 3.2 & 35.0 $\pm$ 1.2 & 31.6 $\pm$ 1.2 & 31.6 $\pm$ 5.3 & \fbox{\textbf{41.0 $\pm$ 2.1}} \\
    & Avg.\ step distance $\downarrow$ & 6.6 $\pm$ 0.7  & 7.0 $\pm$ 0.7  & 6.7 $\pm$ 0.2  & 6.7 $\pm$ 0.6  & 6.8 $\pm$ 1.0  & \fbox{\textbf{6.1 $\pm$ 0.2}} \\
  \midrule
  \multirow{3}{*}{Flash}
    & Step-index acc. (\%) $\uparrow$  & 61.9 $\pm$ 4.8 & 60.3 $\pm$ 3.6 & 65.9 $\pm$ 5.0 & 63.5 $\pm$ 5.5 & 60.3 $\pm$ 3.6 & \fbox{\textbf{69.8 $\pm$ 1.4}} \\
    & Category acc. (\%) $\uparrow$    & 50.0 $\pm$ 6.3 & 48.4 $\pm$ 3.6 & 50.0 $\pm$ 4.8 & 53.2 $\pm$ 5.0 & 54.8 $\pm$ 4.1 & \fbox{\textbf{55.6 $\pm$ 1.4}} \\
    & Avg.\ step distance $\downarrow$ & 0.71 $\pm$ 0.27 & 0.52 $\pm$ 0.06 & 0.52 $\pm$ 0.10 & 0.57 $\pm$ 0.13 & 0.67 $\pm$ 0.16 & \fbox{\textbf{0.44 $\pm$ 0.04}} \\
  \midrule
  \multirow{3}{*}{RelWork}
    & Step-index acc. (\%) $\uparrow$  & 52.6 $\pm$ 3.8 & 48.9 $\pm$ 1.8 & \fbox{\textbf{54.8 $\pm$ 4.2}} & 49.6 $\pm$ 1.0 & 45.9 $\pm$ 4.2 & \fbox{\textbf{54.8 $\pm$ 2.8}} \\
    & Category acc. (\%) $\uparrow$    & \fbox{\textbf{52.6 $\pm$ 4.2}} & 45.9 $\pm$ 4.2 & 44.4 $\pm$ 5.4 & 45.9 $\pm$ 2.8 & 43.7 $\pm$ 5.8 & 46.7 $\pm$ 1.8 \\
    & Avg.\ step distance $\downarrow$ & 2.6 $\pm$ 0.1  & \fbox{\textbf{2.1 $\pm$ 0.5}}  & 2.7 $\pm$ 0.4  & 2.4 $\pm$ 0.1  & 3.2 $\pm$ 0.4  & 2.5 $\pm$ 0.1  \\
    \midrule
  \multirow{3}{*}{Magentic}
    & Step-index acc. (\%) $\uparrow$  & \fbox{\textbf{34.1 $\pm$ 0.0}} & 26.5 $\pm$ 2.8 & 33.3 $\pm$ 2.8 & 31.1 $\pm$ 1.1 & 31.1 $\pm$ 2.1 & 33.3 $\pm$ 1.1 \\
    & Category acc. (\%) $\uparrow$    & 43.2 $\pm$ 1.9 & \fbox{\textbf{46.2 $\pm$ 3.9}} & 43.9 $\pm$ 2.1 & 41.7 $\pm$ 3.9 & 43.9 $\pm$ 4.7 & 41.7 $\pm$ 2.8 \\
    & Avg.\ step distance $\downarrow$ & 19.7 $\pm$ 2.1 & \fbox{\textbf{15.4 $\pm$ 1.9}} & 19.4 $\pm$ 1.9 & 15.5 $\pm$ 0.7 & 22.9 $\pm$ 0.7 & 21.8 $\pm$ 2.2 \\

  \bottomrule
  \end{tabular}%
  }
 \end{table*}


\begin{table}[t]
  \centering
  \caption{\textbf{Failure-category accuracy with DeepSeek-V3.2} as the model for both constraint generation and judging. The two arms use the same judge prompt and differ only in whether \tool's violation log is attached. Mean $\pm$ std over $n{=}3$ judge runs. DeepSeek-V3.2 runs the entire generation stage and produces substantive, checkable constraints in every domain (e.g., 64 violations on Flash, 356 on $\tau$-bench, 148 on RelWork). Constraints improve category accuracy consistently on Flash and $\tau$-bench, with gains of +5.6 to +8.7 points, while results on Magentic-One and RelWork are mixed and broadly neutral. These results show that AgentRx is not GPT-5-specific: the complete pipeline runs unchanged with an open-source model from a different family, and constraint evidence still provides clear gains.}
  \label{tab:deepseek}
  \small
  \setlength{\tabcolsep}{4pt}
  \begin{tabular}{llccr}
  \toprule
  \textbf{Domain} & \textbf{Judge} & \textbf{w/o Vio.} & \textbf{+Vio.} & $\Delta$ \\
  \midrule
  \multirow{2}{*}{Flash}        & Baseline  & 35.7 $\pm$ 1.9 & \textbf{41.3 $\pm$ 2.2} & +5.6 \\
                                & Checklist & 27.8 $\pm$ 3.0 & \textbf{36.5 $\pm$ 4.0} & +8.7 \\
  \midrule
  \multirow{2}{*}{$\tau$-bench} & Baseline  & 21.4 $\pm$ 6.7 & \textbf{28.2 $\pm$ 2.1} & +6.8 \\
                                & Checklist & 20.9 $\pm$ 5.8 & \textbf{27.8 $\pm$ 4.0} & +6.8 \\
  \midrule
  \multirow{2}{*}{Magentic}     & Baseline  & 36.0 $\pm$ 4.0 & 34.7 $\pm$ 4.4 & $-$1.3 \\
                                & Checklist & 27.8 $\pm$ 4.5 & 25.5 $\pm$ 2.5 & $-$2.3 \\
  \midrule
  \multirow{2}{*}{RelWork}      & Baseline  & 30.4 $\pm$ 6.4 & 29.6 $\pm$ 7.3 & $-$0.7 \\
                                & Checklist & 29.6 $\pm$ 2.8 & \textbf{32.6 $\pm$ 3.8} & +3.0 \\
  \bottomrule
  \end{tabular}
\end{table}

\section{Hierarchical Judging Protocols}
\label{app:hierarchical}

\begin{table*}[t]
  \centering
  \caption{\textbf{Two-stage judging protocols vs.\ All-at-Once.} Step-then-Category numbers are the step-by-step constraint-generation ablation from \autoref{tab:ablations}; All-at-Once is the best \tool variant from \autoref{tab:ablations}. Mean $\pm$ std over $n{=}3$ runs with the same \textsc{gpt-5} judge; best per row in a box. In both two-stage protocols the violation log enters at the second (categorization) phase.}
  \label{tab:hierarchical}
  \resizebox{\textwidth}{!}{%
  \begin{tabular}{ll cc cc c}
  \toprule
   & & \multicolumn{2}{c}{\textbf{Step-then-Category}} & \multicolumn{2}{c}{\textbf{Category-then-Step}} & \textbf{All-at-Once} \\
  \cmidrule(lr){3-4} \cmidrule(lr){5-6} \cmidrule(lr){7-7}
  Domain & Metric ($n{=}3$) & w/o Vio. & +Vio. & w/o Vio. & +Vio. & (best) \\
  \midrule
  \multirow{2}{*}{$\tau$-bench}
    & Step acc.\ (\%) $\uparrow$     & 31.6 $\pm$ 1.5 & 38.5 $\pm$ 2.6 & 39.3 $\pm$ 1.2 & 37.6 $\pm$ 1.2 & \fbox{\textbf{42.7 $\pm$ 1.5}} \\
    & Category acc.\ (\%) $\uparrow$ & 32.5 $\pm$ 1.5 & \fbox{\textbf{47.0 $\pm$ 1.5}} & 41.9 $\pm$ 3.2 & 40.2 $\pm$ 4.4 & \fbox{\textbf{47.0 $\pm$ 1.5}} \\
  \midrule
  \multirow{2}{*}{Flash}
    & Step acc.\ (\%) $\uparrow$     & 76.2 $\pm$ 4.1 & 75.4 $\pm$ 6.0 & 85.0 $\pm$ 3.5 & \fbox{\textbf{88.3 $\pm$ 3.1}} & 82.5 $\pm$ 1.4 \\
    & Category acc.\ (\%) $\uparrow$ & 58.7 $\pm$ 1.4 & \fbox{\textbf{65.9 $\pm$ 5.0}} & 62.5 $\pm$ 3.5 & 60.0 $\pm$ 2.0 & \fbox{\textbf{65.9 $\pm$ 5.0}} \\
  \midrule
  \multirow{2}{*}{RelWork}
    & Step acc.\ (\%) $\uparrow$     & 52.6 $\pm$ 1.3 & 54.5 $\pm$ 3.2 & 39.3 $\pm$ 2.1 & 42.2 $\pm$ 0.0 & \fbox{\textbf{61.5 $\pm$ 4.2}} \\
    & Category acc.\ (\%) $\uparrow$ & 48.1 $\pm$ 2.6 & 49.2 $\pm$ 2.8 & 48.1 $\pm$ 3.8 & 47.4 $\pm$ 4.6 & \fbox{\textbf{59.3 $\pm$ 2.1}} \\
  \midrule
  \multirow{2}{*}{Magentic}
    & Step acc.\ (\%) $\uparrow$     & 23.5 $\pm$ 1.3 & 34.1 $\pm$ 0.0 & 26.5 $\pm$ 2.8 & 25.0 $\pm$ 1.9 & \fbox{\textbf{35.6 $\pm$ 1.3}} \\
    & Category acc.\ (\%) $\uparrow$ & 38.6 $\pm$ 0.0 & \fbox{\textbf{46.2 $\pm$ 1.3}} & 31.8 $\pm$ 5.6 & 30.3 $\pm$ 3.9 & \fbox{\textbf{46.2 $\pm$ 1.3}} \\
  \bottomrule
  \end{tabular}%
  }
\end{table*}

\tool's default All-at-Once judge predicts the critical step and category jointly.
\autoref{tab:ablations} also evaluates Step-then-Category, which localizes the step first and then classifies it.
Here we add the inverse ordering, Category-then-Step: the judge first infers the failure category from the full trajectory, then localizes the failure step consistent with that category.
\autoref{tab:hierarchical} compares all three protocols with and without \tool's violation evidence.
Both two-stage orderings suffer from error propagation: committing to a category (or a step) before the joint pass discards grounding that a single pass can use, and any first-stage error constrains the second stage.
Category-then-Step outperforms Step-then-Category on $\tau$-bench and Flash but underperforms on RelWork and Magentic, where long, nested traces make an early category commitment especially costly.
The one case where decomposition beats the joint judge is Flash step localization (88.3\% vs.\ 82.5\%), on the domain with the shortest trajectories.
Everywhere else All-at-Once matches or exceeds both two-stage variants, so it remains the default protocol; two-stage decomposition offers only selective gains.

\section{Root-Cause Attribution VS First Failure}
\label{app:rootcause}
The following trajectory demonstrates why root-cause failure attribution is the better objective for evaluating agent reliability. 
The Who\&When dataset labels step 3 as failure citing that ``WebSurfer's inability to reliably access the requested documents resulted in the overall task failure, as the necessary time span data could not be extracted or compared. This underscores the need for enhanced fallback mechanisms and more robust search strategies."
This diagnosis misses a key fact: the agent recovers from this early access failure, so it is not the cause of the eventual outcome.
We annotate the key failure as shown below:

\begin{minted}{json}
{
"trajectory_id": "5f982798-16b9-4051-ab57-cfc7ebdb2a91",
"failures": [
{
    "failure_id": 1,
    "step_number": 13,
    "step_reason": "Websurfer could not  download a PDF file and search throught it which was an instruction given by Orchestrator.",
    "failure_category": "Instruction/Plan Adherence Failure",
    "category_reason": "Instruction not followed, the agent did not download and search through the PDF file as instructed",
    "failed_agent": "Websurfer"
},
{
    "failure_id": 2,
    "step_number": 17,
    "step_reason": "Websurfer could not  download a PDF file and search throught it which was an instruction given by Orchestrator",
    "failure_category": "Instruction/Plan Adherence Failure",
    "category_reason": "Websurfer could not  download a PDF file and search throught it which was an instruction given by Orchestrator",
    "failed_agent": "Websurfer"
},
{
    "failure_id": 3,
    "step_number": 33,
    "step_reason": "FileSurfer hallucinated. It downloaded the file at '/workspace/workspace/http:/export.arxiv.org/pdf/2007.xx' but later attempted to read a non-existent file 'file:///workspace/path_to_july_2020_paper.pdf'.",
    "failure_category": "Invention of new information",
    "category_reason": "FileSurfer hallucinated. It downloaded the file at '/workspace/workspace/http:/export.arxiv.org/pdf/2007.xx' but later attempted to read a non-existent file 'file:///workspace/path_to_july_2020_paper.pdf'.",
    "failed_agent": "FileSurfer"
}
],
"root_cause": {
"failure_id": 3,
"reason_for_root_cause": "The Orchestrator was able to recover from earlier errors but the FileSurfer hallucination was a critical failure that prevented further progress."
},
"failure_summary": "The agent could not download and read the specified PDF file due to a hallucination by the FileSurfer agent."
}
\end{minted}

Failures \#1–\#2 were recoverable because the orchestrator successfully re-routed execution when WebSurfer stalled. It explicitly acknowledged missing information and switched to FileSurfer, whose file-centric capabilities are better suited for handling PDFs. In other words, the system demonstrated a working fallback: early retrieval failures triggered tool substitution and continued progress rather than terminal collapse.
Failure \#3 was the key failure because FileSurfer saved file to the wrong path and subsequent operations were based on non-existent data, making recovery impossible.


 \begin{figure}[t]
 \centering
 \includegraphics[width=\columnwidth]{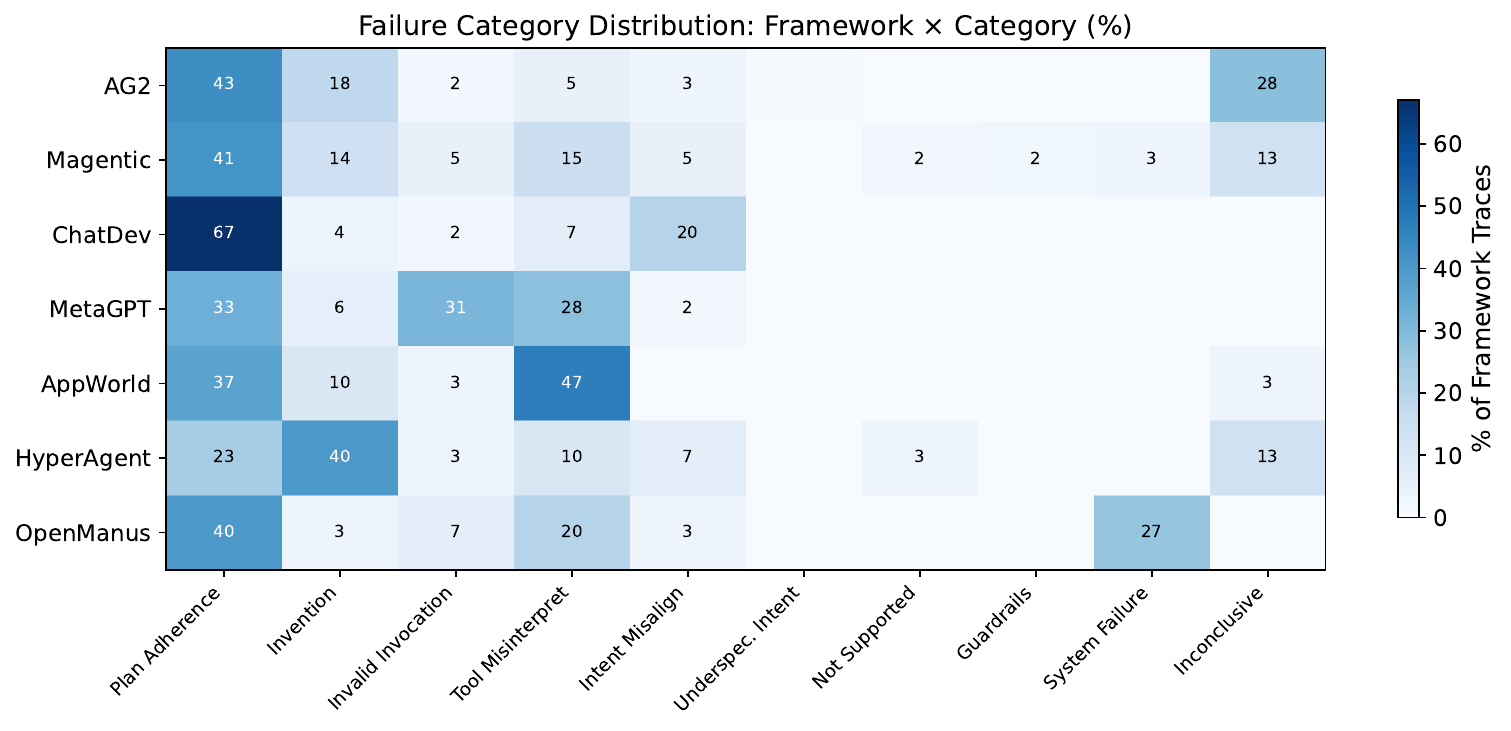}
 \caption{Failure category distribution (\%) per framework on 1,022 MAST trajectories. Each row is normalized independently.}
 \label{fig:mast-heatmap}
 \end{figure}
 
\begin{figure}[t]
\centering
\includegraphics[width=\columnwidth]{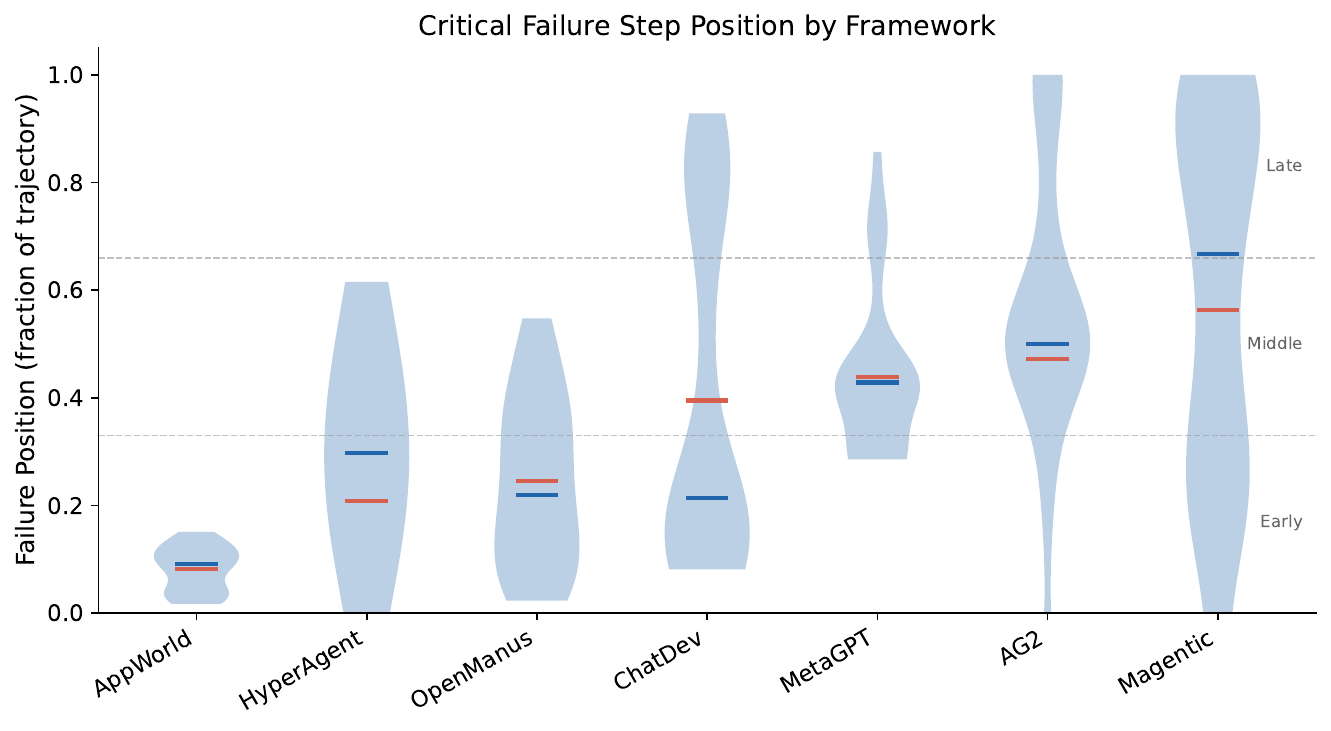}
\caption{Distribution of critical failure step position (as fraction of trajectory length) by framework in the entire 1022 MAST dataset. Red marks indicate means; blue marks indicate medians.}
\label{fig:mast-violin}
\end{figure}

\section{Failure Annotation for Magentic One}
\label{app:magentic-22}

An example annotation for a magentic trajectory in our benchmark. We annotate a total of 22 failures in this trajectory as shown:

\begin{minted}{json}
{
"trajectory_id": "16d825ff-1623-4176-a5b5-42e0f5c2b0ac",
"failures": [
{
    "failure_id": 1,
    "step_number": 5,
    "step_reason": "WebSurfer did not get arrival time information which is important",
    "failure_category": "Instruction/Plan Adherence Failure",
    "category_reason": "WebSurfer did not get arrival time information which is important",
    "failed_agent": "WebSurfer"
},
{
    "failure_id": 2,
    "step_number": 13,
    "step_reason": "WebSurfer did  scroll but did not get information for specified date",
    "failure_category": "Instruction/Plan Adherence Failure",
    "category_reason": "WebSurfer did  scroll but did not get information for specified date",
"failed_agent": "WebSurfer"
},
{
    "failure_id": 3,
    "step_number": 17,
    "step_reason": "WebSurfer did not get information for specified date",
    "failure_category": "Instruction/Plan Adherence Failure",
    "category_reason": "WebSurfer did not get information for specified date",
"failed_agent": "WebSurfer"
},
{
"failure_id": 4,
"step_number": 25,
"step_reason": "WebSurfer did not get information for specified date",
"failure_category": "Instruction/Plan Adherence Failure",
"category_reason": "WebSurfer did not get information for specified date",
"failed_agent": "WebSurfer"
},
{
    "failure_id": 5,
    "step_number": 29,
    "step_reason": "WebSurfer did not get information for specified date",
    "failure_category": "Instruction/Plan Adherence Failure",
    "category_reason": "WebSurfer did not get information for specified date",
    "failed_agent": "WebSurfer"
},
{
"failure_id": 6,
"step_number": 33,
"step_reason": "WebSurfer did not get information for specified date",
"failure_category": "Instruction/Plan Adherence Failure",
"category_reason": "WebSurfer did not get information for specified date",
"failed_agent": "WebSurfer"
},
{
    "failure_id": 7,
    "step_number": 37,
    "step_reason": "WebSurfer did not get information for specified date",
    "failure_category": "Instruction/Plan Adherence Failure",
    "category_reason": "WebSurfer did not get information for specified date",
    "failed_agent": "WebSurfer"
},
{
"failure_id": 8,
"step_number": 41,
"step_reason": "WebSurfer did not get information for specified date",
"failure_category": "Instruction/Plan Adherence Failure",
"category_reason": "WebSurfer did not get information for specified date",
"failed_agent": "WebSurfer"
},
{
    "failure_id": 9,
    "step_number": 56,
    "step_reason": "WebSurfer did not get information for specified date",
    "failure_category": "Instruction/Plan Adherence Failure",
    "category_reason": "WebSurfer did not get information for specified date",
    "failed_agent": "WebSurfer"
},
{
    "failure_id": 10,
    "step_number": 60,
    "step_reason": "WebSurfer did not get information for specified date",
    "failure_category": "Instruction/Plan Adherence Failure",
    "category_reason": "WebSurfer did not get information for specified date",
    "failed_agent": "WebSurfer"
},
{
    "failure_id": 10,
    "step_number": 60,
    "step_reason": "WebSurfer did not get information for specified date",
    "failure_category": "Instruction/Plan Adherence Failure",
    "category_reason": "WebSurfer did not get information for specified date",
    "failed_agent": "WebSurfer"
},
{
    "failure_id": 11,
    "step_number": 66,
    "step_reason": "Orchestrator try to contact through email which might not be good strategy",
    "failure_category": "Intent Plan Misalignment",
    "category_reason": "Orchestrator try to contact through email which might not be good strategy",
    "failed_agent": "Orchestrator"
},
{
    "failure_id": 12,
    "step_number": 70,
    "step_reason": "Orchestrator did not properly interpret user intent again trying to email which is not good strategy",
    "failure_category": "Intent Plan Misalignment",
    "category_reason": "Orchestrator did not properly interpret user intent again trying to email which is not good strategy",
    "failed_agent": "Orchestrator"
},
{
    "failure_id": 13,
    "step_number": 74,
    "step_reason": "Orchestrator did not properly interpret user intent again trying to email which is not good strategy",
    "failure_category": "Intent Plan Misalignment",
    "category_reason": "Orchestrator did not properly interpret user intent again trying to email which is not good strategy",
    "failed_agent": "Orchestrator"
},
{
    "failure_id": 14,
    "step_number": 78,
    "step_reason": "Orchestrator did not properly interpret user intent again trying to email which is not good strategy",
    "failure_category": "Intent Plan Misalignment",
    "category_reason": "Orchestrator did not properly interpret user intent again trying to email which is not good strategy",
    "failed_agent": "Orchestrator"
},
{
    "failure_id": 15,
    "step_number": 82,
    "step_reason": "Orchestrator did not properly interpret user intent again trying to email which is not good strategy",
    "failure_category": "Intent Plan Misalignment",
    "category_reason": "Orchestrator did not properly interpret user intent again trying to email which is not good strategy",
    "failed_agent": "Orchestrator"
},
{
    "failure_id": 16,
    "step_number": 86,
    "step_reason": "Orchestrator did not properly interpret user intent again trying to email which is not good strategy",
    "failure_category": "Intent Plan Misalignment",
    "category_reason": "Orchestrator did not properly interpret user intent again trying to email which is not good strategy",
    "failed_agent": "Orchestrator"
},
{
    "failure_id": 17,
    "step_number": 90,
    "step_reason": "Orchestrator did not properly interpret user intent again trying to email which is not good strategy",
    "failure_category": "Intent Plan Misalignment",
    "category_reason": "Orchestrator did not properly interpret user intent again trying to email which is not good strategy",
    "failed_agent": "Orchestrator"
},
{
    "failure_id": 18,
    "step_number": 101,
    "step_reason": "Orchestrator did not properly interpret user intent again trying to email which is not good strategy",
    "failure_category": "Intent Plan Misalignment",
    "category_reason": "Orchestrator did not properly interpret user intent again trying to email which is not good strategy",
    "failed_agent": "Orchestrator"
},
{
    "failure_id": 19,
    "step_number": 105,
    "step_reason": "Orchestrator did not properly interpret user intent again trying to email which is not good strategy",
    "failure_category": "Intent Plan Misalignment",
    "category_reason": "Orchestrator did not properly interpret user intent again trying to email which is not good strategy",
    "failed_agent": "Orchestrator"
},
{
    "failure_id": 20,
    "step_number": 109,
    "step_reason": "Orchestrator did not properly interpret user intent again trying to email which is not good strategy",
    "failure_category": "Intent Plan Misalignment",
    "category_reason": "Orchestrator did not properly interpret user intent again trying to email which is not good strategy",
    "failed_agent": "Orchestrator"
},
{
    "failure_id": 21,
    "step_number": 116,
    "step_reason": "Orchestrator did not properly interpret user intent again trying to email which is not good strategy",
    "failure_category": "Intent Plan Misalignment",
    "category_reason": "Orchestrator did not properly interpret user intent again trying to email which is not good strategy",
    "failed_agent": "Orchestrator"
},
{
    "failure_id": 22,
    "step_number": 120,
    "step_reason": "Orchestrator did not properly interpret user intent again trying to email/phone which is not good strategy",
    "failure_category": "Intent Plan Misalignment",
    "category_reason": "Orchestrator did not properly interpret user intent again trying to email/phone which is not good strategy",
    "failed_agent": "Orchestrator"
}
"root_cause": {
"failure_id": 1,
"reason_for_root_cause": "The primary root cause of the failures in this trajectory is the WebSurfer's inability to retrieve specific arrival time information for the specified date. This lack of crucial information led to a series of misinterpretations by the Orchestrator, which repeatedly attempted to contact through email a strategy misaligned with the user's intent. The cascading effect of these initial shortcomings resulted in multiple failures throughout the trajectory."
}
}
\end{minted}

\section{Agent Trajectories for each category of our failure taxonomy across domains}

\subsection{System Failure (Flash)}
\begin{minted}{json}
{
"trajectory_id": "9_withouths_tip_session_2_417931231",
"failures": [
{
"failure_id": 1,
"step_number": 3,
"step_reason": "KustoApiError: Error getting schema for Cluster='https://azcore1.southeastasia.kusto.windows.net/': Failed to connect to the remote cluster: InternalServiceError (520-UnknownError) followed by a SyntaxError of the Kusto query",
"failure_category": "System Failure",
"category_reason": "Connection failure error, system error + syntax error",
"failed_agent": "KustoAgent"
}
],
"root_cause": {
    "failure_id": 1,
    "reason_for_root_cause": "Connection failure error, system error + syntax error"
    },
    "failure_summary": "System failure + Syntax errors"
}    
\end{minted}
\subsection{Misinterpretation of Tool Output ($\tau$-bench)}
\begin{minted}{json}

"trajectory_id": 2,
"failures": [
{
"failure_id": 1,
"step_number": 3,
"step_reason": "At step 3, the assistant agent did not authenticate user information before proceeding to provide information about available t-shirts",
"failure_category": "Instruction Adherence Failure",
"category_reason": "The assistant agent did not follow the expected policy of authenticating user information before providing product details.",
"failed_agent": "Assistant"
},
{
"failure_id": 2,
"step_number": 7,
"step_reason": "At step 7, the agent did not correctly count the number of available t-shirts from the tool call result.",
"failure_category": "Misinterpretation of Tool Output",
"category_reason": "The assistant misinterpreted the output from the tool call, leading to an incorrect count of available t-shirts.",
"failed_agent": "Assistant"
  }
],
"root_cause": {
  "failure_id": 2,
  "reason_for_root_cause": "The assistant finally did authenticate before providing user specific information. The incorrect count does not correspond with ground truth output."
},
"failure_summary": "The agent did not correctly count the number of available t-shirts from the tool call result."
  }    
\end{minted}
\subsection{Instruction Adherence Failure (Flash)}
\begin{minted}{json}
{
"trajectory_id": "7_withhs_tip_session_2_424614956",
"failures": [
{
"failure_id": 1,
"step_number": 4,
"step_reason": "The actual solution is to \"Delete the VM through the provided link, or contact the resource owner to delete it.\" The model's answer does not explicitly suggest deleting the VM or contacting the resource owner to do so. However, it does guide the user to manually inspect and clean up any lingering VMs or resources, which partially aligns with the intent of deleting the resource. The model fails to mention using a provided link or directly contacting the resource owner, which are key steps in the actual solution. Thus, while there is some overlap specifically the suggestion to clean up resources the solution is incomplete and misses key instructions.",
"failure_category": "Instruction/Plan Adherence Failure",
"category_reason": "incomplete/absent conclusion/mitigation step and also did not provide the Azure link.",
"failed_agent": "Orchestrator"
}
],
"root_cause": {
"failure_id": 1,
"reason_for_root_cause": "incomplete/absent conclusion/mitigation step and also did not provide the Azure home link"
},
"failure_summary": "The model's answer does not follow the plan perfectly"
}
\end{minted}
\subsection{Invention of New Information (Flash)}
\begin{minted}{json}
{
"trajectory_id": "9_withouths_tip_session_1_445308210",
"failures": [
{
"failure_id": 1,
"step_number": 3,
"step_reason": "Step 3 incorrect query, even though the Kusto query returned None; the agent tried with a new invented python script.",
"failure_category": "Invention of New Information",
"category_reason": "hallucination of python script",
"failed_agent": "Coder"
},
{
"failure_id": 2,
"step_number": 5,
"step_reason": "The GeneralAssistant came up with a link https://portal.azure.com/#search/152076538 instead of providing the home page as per the plan.",
"failure_category": "Invention of New Information",
"category_reason": "hallucination of link",
"failed_agent": "GeneralAssistant"
}
],
"root_cause": {
  "failure_id": 1,
  "reason_for_root_cause": "hallucination of python script + link"
},
"failure_summary": "hallucination, extra steps executed"
}    
\end{minted}
\subsection{Intent Not Supported (Magentic)}
\begin{minted}{json}
{
"trajectory_id": "a1e91b78-d3d8-4675-bb8d-62741b4b68a6",
"failures": [
{
"failure_id": 1,
"step_number": 5,
"step_reason": "Websurfer was asked to take snapshot in youtube video but it could not do such actions",
"failure_category": "Intent not supported",
"category_reason": "Websurfer was asked to take snapshot in youtube video but it could not do such actions",
"failed_agent": "Websurfer"
},
{
"failure_id": 2,
"step_number": 9,
"step_reason": "Websurfer was asked to take snapshot in youtube video but it could not do such actions",
"failure_category": "Intent not supported",
"category_reason": "Websurfer was asked to take snapshot in youtube video but it could not do such actions",
"failed_agent": "Websurfer"
},
{
"failure_id": 3,
"step_number": 13,
"step_reason": "Websurfer was asked to take snapshot in youtube video but it could not do such actions",
"failure_category": "Intent not supported",
"category_reason": "Websurfer was asked to take snapshot in youtube video but it could not do such actions",
"failed_agent": "Websurfer"
},
{
"failure_id": 4,
"step_number": 17,
"step_reason": "Websurfer was asked to take snapshot in youtube video but it could not do such actions",
"failure_category": "Intent not supported",
"category_reason": "Websurfer was asked to take snapshot in youtube video but it could not do such actions",
"failed_agent": "Websurfer"
},
{
"failure_id": 5,
"step_number": 21,
"step_reason": "Websurfer was asked to take snapshot in youtube video but it could not do such actions",
"failure_category": "Intent not supported",
"category_reason": "Websurfer was asked to take snapshot in youtube video but it could not do such actions",
"failed_agent": "Websurfer"
},
{
"failure_id": 6,
"step_number": 25,
"step_reason": "Websurfer was asked to take snapshot in youtube video but it could not do such actions",
"failure_category": "Intent not supported",
"category_reason": "Websurfer was asked to take snapshot in youtube video but it could not do such actions",
"failed_agent": "Websurfer"
}
],
"root_cause": {
"failure_id": 1,
"reason_for_root_cause": "The first error occurred when the Websurfer agent was asked to take a snapshot in a YouTube video, which is beyond its capabilities. Subsequent failures were repetitions of this unsupported intent."
},
"failure_summary": "The Websurfer agent was repeatedly asked to perform an action (taking snapshots in a YouTube video) that it is not capable of executing, leading to failure."
   
\end{minted}

\subsection{Underspecified User Intent ($\tau$-bench)}
\begin{minted}{json}
{
"trajectory_id": 71,
"failures": [
{
"failure_id": 26,
"step_number": 24,
"step_reason": "At step 24, the task instruction did not specify the type of black lamp and since there are multiple black lamps available, it wont be possible to match the ground truth actions.",
"failure_category": "Underspecified User Intent",
"category_reason": "The task instruction is underspecified regarding the type of black lamp to be added to the cart.",
"failed_agent": "User"
},
{
"failure_id": 27,
"step_number": 27,
"step_reason": "Assistant prematurely called modify items tool on a pending order which locked the order and the user later on wasn't able to change the backpack that he wanted.",
"failure_category": "Intent-Plan Misalignment",
"category_reason": "Here, the plan generated by the assistant is incorrect, it should not have prematurely called modify items tool on a pending order before finalizing all the items to be modified in the user's order.",
"failed_agent": "Assistant"
}
],
"root_cause": {
"failure_id": 26,
"reason_for_root_cause": "The task instruction did not specify the type of black lamp and since there are multiple black lamps available, it led to incorrect matching of the ground truth actions. Did not recover from the error."
},
"failure_summary": "The task instruction did not specify the type of black lamp and since there are multiple black lamps available, it wont be possible to match the ground truth actions."
  }
\end{minted}

\subsection{Guardrails Triggered (Magentic)}
\begin{minted}{json}
{
 "task_instruction": "You are a business analyst evaluating cloud storage solutions. You need to compare the current pricing plans for DataVault's Pro and Enterprise tiers to make an informed purchasing decision for your company. Request the agent to find and summarize the pricing information including features and costs for both tiers.",
 "trajectory_snippet": [
{
    "content": "Can you find the current pricing plans for the Pro and Enterprise tiers on the DataVault cloud storage service? I need to compare them for my business decision.",
    "role": "human",
    "index": 1
},
{
    "content": "I'll help you find that pricing information. Let me navigate to DataVault's website and locate their pricing details.",
    "role": "Orchestrator",
    "index": 2
},
{
    "content": "Please navigate to DataVault's official website, find the pricing page, and extract the pricing and features for both Pro and Enterprise tiers.",
    "role": "Orchestrator (-> WebSurfer)",
    "index": 3
},
{
    "content": "I navigated to 'https://www.datavault.io/pricing'.\n\n**Detected blocking mechanism:**\n\nThe page displays:\n\n\"Access Denied\n\nYou do not have permission to access this page.\n\nReason: Our system has detected that your request appears to be coming from an automated bot or script.\n\nTo protect our users and prevent abuse, we require human verification for pricing information.\n\nError Code: 403 - Forbidden (Bot Detection)\nReference ID: CDN-BLK-98472\n\nIf you are a human user, please:\n1. Enable JavaScript in your browser\n2. Complete the CAPTCHA challenge below\n3. Contact support@datavault.io if this problem persists\n\n[CAPTCHA Challenge - Please verify you are human]\n\nThis page is protected by Cloudflare Bot Management.\"\n\n**Navigation blocked:** The website has detected automated access and requires human verification (CAPTCHA) to proceed. I cannot bypass this security measure as it violates the site's terms of service and my operational guidelines.\n\nI am unable to access the pricing information due to the bot detection and CAPTCHA requirement.",
    "role": "WebSurfer",
    "index": 4
},
{
    "content": "I see the direct access is blocked. Let me try searching for the pricing information from alternative sources or cached content.",
    "role": "Orchestrator",
    "index": 5
},
{
    "content": "Please search for 'DataVault Pro Enterprise pricing comparison' or try to find cached or third-party sources that list this pricing information.",
    "role": "Orchestrator (-> WebSurfer)",
    "index": 6
},
{
    "content": "I performed a search for 'DataVault Pro Enterprise pricing comparison'.\n\nThe search results show several third-party review sites and comparison pages, but when attempting to access detailed pricing information, I encountered similar restrictions:\n\n1. **SoftwareReviews.com** - Pricing details require login\n2. **TechCompare.io** - Shows \"Contact for pricing\" with no specific numbers\n3. **G2.com** - Lists features but pricing marked as \"Available upon request\"\n4. **Cached pages** - Pricing information is outdated (from 2022) and may not be current\n\nUnfortunately, DataVault appears to keep their Enterprise tier pricing behind access controls across multiple platforms. Most sources either require account registration/login, display \"Contact sales\" instead of actual prices, or have outdated cached information.\n\nI cannot retrieve the current, accurate pricing information the user needs due to these access restrictions.",
    "role": "WebSurfer",
    "index": 7
},
{
    "content": "I apologize, but I was unable to retrieve the current pricing information for DataVault's Pro and Enterprise tiers. The DataVault website has bot detection and CAPTCHA protection that prevents automated access to their pricing page (Error Code: 403 - Forbidden). Additionally, alternative sources like review sites and comparison platforms either require login credentials, show \"Contact sales for pricing,\" or contain outdated information.\n\nTo get this information, you would need to:\n1. Visit https://www.datavault.io/pricing directly in your browser\n2. Complete the CAPTCHA verification\n3. Or contact DataVault's sales team at support@datavault.io for Enterprise pricing details\n\nThis limitation is due to website access restrictions, not a capability issue with our search approach.",
    "role": "Assistant",
    "index": 8
}
 ],
    "reason_for_category": "The category is guardrails triggered because at step 4, the WebSurfer agent encountered an explicit access denial when attempting to navigate to DataVault's pricing page. The website returned a 403 Forbidden error with bot detection, requiring CAPTCHA verification to proceed. This is an external site access restriction (Cloudflare Bot Management) that blocks automated agents from accessing the content. The agent's plan was valid and would have worked if this access restriction were removed - it correctly identified the target URL and attempted proper navigation. At step 7, alternative approaches (searching third-party sources, cached pages) also failed due to similar access controls (login requirements, paywalls). This is not a malformed tool invocation (the navigation commands were correct), not an infrastructure failure (the systems are functioning as designed), and not a planning error (the approach was appropriate). The failure is purely due to external guardrails designed to prevent automated access to the pricing information."
} 
\end{minted}

\subsection{Intent Plan Misalignment ($\tau$-bench)}
\begin{minted}{json}
{
"trajectory_id": 28,
"failures": [
{
"failure_id": 13,
"step_number": 33,
"step_reason": "At step 33,  the assistant mistakenly believes that it can cancel a subset of a pending order which is not allowed as per domain policy, as a result the entire order got cancelled instead of just the garden hose.",
"failure_category": "Intent-Plan Misalignment",
"category_reason": "The assistant came up with an incorrect plan based on a wrong assumption that a subset of an order can be cancelled which violates the domain policy.",
"failed_agent": "Assistant"
}
],
"root_cause": {
"failure_id": 13,
"reason_for_root_cause": "The assistant called cancel order on the entire order which led to an incorrect final outcome as compared to ground truth actions."
},
"failure_summary": "Assistant mistakenly believes that it can cancel a subset of the pending order which is not allowed as per domain policy, as a result the entire order got cancelled instead of just the hose."
}
\end{minted}
\subsection{Invalid Invocation ($\tau$-bench)}
\begin{minted}{json}
{
"trajectory_id": 34,
"failures": [
{
"failure_id": 16,
"step_number": 17,
"step_reason": "At step 17, the assistant uses modify order to cancel a subset of orders, however modify orders also need to have a replacement, which it did not provide resulting in an illegal tool call",
"failure_category": "Invalid Invocation",
"category_reason": "The assistant calls the modify order tool with invalid arguments.",
"failed_agent": "Assistant"
},
{
"failure_id": 17,
"step_number": 21,
"step_reason": "At step 21, the assistant tries to bypass the modify tool argument restriction by trying to modify the item_id with the same item_id in order to try to cancel it",
"failure_category": "Invalid Invocation",
"category_reason": "The assistant again calls modify order tool with invalid arguments trying to bypass the previous error.",
"failed_agent": "Assistant"
},
{
"failure_id": 18,
"step_number": 30,
"step_reason": "At step 30, the user does not ask the assistant to modify the address of the current pending order but is clearly a part of the overall task instruction.",
"failure_category": "Underspecified User Intent",
"category_reason": "The user does not ask the assistant to do all the tasks as mentioned in the task instruction, hence underspecifying its intent.",
"failed_agent": "User"
}
],
"root_cause": {
"failure_id": 16,
"reason_for_root_cause": "The assistant called the modify order tool with invalid arguments, leading to an illegal tool call, and the agent did not recover from this error."
},
"failure_summary": "Assistant used modify order to cancel a subset of orders, but modify order requires a replacement which was not provided - illegal tool call."
  }
\end{minted}
\section{Constraint Generation}
\label{app:constraint-gen}
\subsection{\tool Constraint Schema Output Format}
\label{sec:schema}
\lstdefinelanguage{json}{
  basicstyle=\ttfamily\footnotesize,
  numbers=left,
  numberstyle=\tiny,
  stepnumber=1,
  numbersep=6pt,
  showstringspaces=false,
  breaklines=true,
  frame=single,
  columns=fullflexible,
  literate=
   *{0}{{{\color{black}0}}}{1}
    {1}{{{\color{black}1}}}{1}
    {2}{{{\color{black}2}}}{1}
    {3}{{{\color{black}3}}}{1}
    {4}{{{\color{black}4}}}{1}
    {5}{{{\color{black}5}}}{1}
    {6}{{{\color{black}6}}}{1}
    {7}{{{\color{black}7}}}{1}
    {8}{{{\color{black}8}}}{1}
    {9}{{{\color{black}9}}}{1}
}

\begin{minted}{json}
{
  "assertion_name": "string_unique_snake_case",
  "taxonomy_targets": [
    "Instruction/PlanAdherenceFailure",
    "InventionOfNewInformation",
    "InvalidInvocation",
    "MisinterpretationOfToolOutput",
    "IntentPlanMisalignment",
    "UnderspecifiedUserIntent",
    "IntentNotSupported",
    "GuardrailsTriggered",
    "SystemFailure"
  ],
  "constraint_type": "SCHEMA | PROTOCOL | RELATIONAL_POST | PROVENANCE | TEMPORAL | CAPABILITY | ANY",
  "event_trigger": {
    "step_index": "*|int|range",
    "substep_index": "*|int|range",
    "role_name": "AgentName_or_*",
    "content_regex": "regex_or_*",
    "tool_name": "ToolName_or_*"
  },
  "check_hint": "deterministic procedure description in 2-8 sentences",
  "examples": {
    "pass_scenario": "short",
    "fail_scenario": "short"
  },
  "check_type": "python_check|nl_check",
  "python_check": {
    "function_name": "same_as_assertion_name",
    "args": ["trajectory", "current_step_index"],
    "code_lines": [
      "def same_as_assertion_name(trajectory, current_step_index):",
      "    \"\"\"Return True iff invariant holds.\"\"\"",
      "    # parse trajectory[current_step_index], look back if needed",
      "    # MUST include at least one explicit failure path: return False",
      "    return True"
    ]
  },
  "nl_check": {
    "judge_system_prompt_template": "{NL_CHECK_JUDGE_SYSTEM_PROMPT}",
    "judge_user_prompt_template": "template using {POLICY_TEXT} {CURRENT_EVENT_JSON} {WINDOW_EVENTS_JSON}",
    "judge_scope_notes": "what events are in scope and what counts as evidence",
    "focus_steps_instruction": "REQUIRED: Identify 2-4 specific events by relative position and what to check in each.",
    "judge_rubric": ["objective criterion 1", "objective criterion 2", "..."],
    "rubric_evaluation_algorithm_template": "{RUBRIC_EVALUATION_ALGORITHM}",
    "output_format_template": "{OUTPUT_FORMAT}"
  }
}
\end{minted}

\subsection{Semantic Check generated by \tool}
\begin{minted}{json}
{
"assertion_name": "explicit_user_confirmation_before_write_actions",
"taxonomy_targets": [
  "Instruction/PlanAdherenceFailure",
  "IntentPlanMisalignment"
],
"invariant_type": "TEMPORAL",
"event_trigger": {
  "role_name": "assistant",
  "content_regex": "*",
  "tool_name": "cancel_pending_order|exchange_delivered_order_items|return_delivered_order_items|modify_pending_order_.*|modify_user_address"
},
"check_hint": "Before any write-action tool invocation (cancel, exchange, return, modify order/address), verify that the assistant previously described the intended action and target ID, and that the user explicitly confirmed proceeding (e.g., 'yes', 'confirm', 'please proceed').",
"check_type": "nl_check",
"python_check": {},
"nl_check": {
"judge_system_prompt_template": "You are a strict compliance judge. Evaluate only with evidence in the provided events and the policy text.\nDo not infer intent beyond explicit statements.\nCRITICAL: If required evidence is missing or ambiguous, mark the criterion as UNCLEAR.\nWhen criteria are marked as UNCLEAR, they are discarded and do NOT contribute to a fail verdict.\nIf all criteria are UNCLEAR or CLEAR_PASS after discarding UNCLEAR ones, the final verdict is PASS.\nOnly mark as CLEAR_FAIL when there is definitive, unambiguous evidence of violation.\nAmbiguity or missing evidence leads to PASS, not FAIL.\n",
"judge_user_prompt_template": "POLICY TEXT:\n{POLICY_TEXT}\n\nTASK INSTRUCTION:\n{TASK_INSTRUCTION}\n\nCURRENT EVENT:\n{CURRENT_EVENT_JSON}\n\nCONTEXT WINDOW EVENTS:\n{WINDOW_EVENTS_JSON}\n\nEvaluate whether, before the current write-action tool call (cancel, modify, exchange, return, address update), the assistant clearly described the action and entity ID, the user explicitly confirmed proceeding with that action, and the IDs match between description and the tool call.",
"judge_scope_notes": "Judge only within the window of events provided. Determine whether this specific write-action tool call is properly preceded by a clear assistant explanation and explicit user confirmation for the same action and same order_id/user_id.",
"focus_steps_instruction": "Focus on: (1) The current assistant tool-call step to identify which write-action tool is being invoked and what identifier (order_id or user_id) is present in its arguments. (2) The immediately prior user message to check for explicit affirmation language indicating consent to proceed with the described action. (3) Assistant messages in the 2-3 steps before the current tool call to see whether the assistant described the intended action (cancel/modify/exchange/return/address update), reminded the user to confirm all relevant items (for exchange/modify-items), and mentioned the same identifier as in the tool call.",
"judge_rubric": [
  "There exists an assistant message earlier in the context that explicitly describes the specific write action type (cancel, modify, exchange, return, or address update) and includes the same identifier (order_id or user_id) that appears in the current tool call arguments.",
  "There exists a user message after the assistant's action description and before the current tool call that contains explicit confirmation language agreeing to proceed with that specific action (e.g., 'yes', 'confirm', 'please proceed', 'go ahead').",
  "When the action is exchange or modify-items, the assistant reminded the customer to confirm they have provided all items to be changed; and the current tool call scope is consistent with the items described and confirmed."
],
"rubric_evaluation_algorithm_template": "Rubric Evaluation Algorithm:\nStep 1: For each criterion in the rubric, evaluate whether it can be CLEARLY judged as PASS or FAIL based solely on the provided events.\n- Mark as CLEAR_PASS if the criterion is demonstrably satisfied by the evidence\n- Mark as CLEAR_FAIL if the criterion is demonstrably violated by the evidence  \n- Mark as UNCLEAR if: insufficient events to judge, criterion is ambiguous in this context, or pass/fail cannot be decisively determined\n\nStep 2: Discard ALL criteria marked as UNCLEAR from consideration.\n\nStep 3: Determine final verdict:\n- If ANY remaining criterion is CLEAR_FAIL == return verdict 'fail'\n- If ALL remaining criteria are CLEAR_PASS (or no criteria remain after discarding UNCLEAR) == return verdict 'pass'\n\nImportant: Only fail when you have CLEAR evidence of failure. When in doubt, mark as UNCLEAR and discard.\n",
"output_format_template": "Output Format:\nYou MUST return a JSON response with this EXACT structure:\n{\n  \"verdict\": \"pass\" or \"fail\",\n  \"rubric_results\": [\n    {\n      \"criterion_index\": 0,\n      \"criterion\": \"full text of the criterion from the rubric\",\n      \"evaluation\": \"CLEAR_PASS\" or \"CLEAR_FAIL\" or \"UNCLEAR\",\n      \"reasoning\": \"detailed explanation of why this evaluation was chosen, citing specific evidence from events\"\n    }\n    // ... one entry per rubric criterion\n  ],\n  \"final_reasoning\": \"explanation of how the verdict was determined from rubric_results, stating which criteria were discarded as UNCLEAR and which criteria drove the final decision\"\n}\n"
}
}
\end{minted}

\section{\tool Violation Examples}

\tool generates the following programmatic violation because the LLM-extracted “number of t-shirts” from the tool output does not match the true t-shirt count, so the check flags a mismatch.

\begin{minted}{json}
{
    "assertion_name": "tshirt_available_options_match_variants_count",
    "taxonomy_targets": [
        "MisinterpretationOfToolOutput",
    ],
    "invariant_type": "RELATIONAL_POST",
    "event_trigger": {
        "step_index": 7,
        "role_name": "assistant",
        "content_regex": "T-?shirt|t-?shirt|T-Shirt",
        "tool_name": "*"
    },
    "check_hint": "When the assistant now reports how many T-shirt options are available, it should compute the count from the variants field of the latest get_product_details result for the T-Shirt product. Specifically, count how many variant entries have available == true and verify that this equals the numeric count stated by the assistant for 'available' T-shirt options. If the assistant gives an 'available options' count that does not match this filtered count, or gives any numeric options count without a corresponding get_product_details result, the invariant fails.",
    "check_type": "python_check",
    "python_check": {
    "function_name": "tshirt_available_options_match_variants_count",
    "args": [
    "trajectory",
    "current_step_index"
    ],
    "code_lines": [
"import re",
"import json",
"",
"def tshirt_available_options_match_variants_count(trajectory, current_step_index):",
"    '''Verify that the assistant's stated count of available T-shirt options matches the tool response.'''",
"    print('Function: tshirt_available_options_match_variants_count')",
"    ",
"    # Access the current step from the trajectory IR format",
"    steps = trajectory.get('steps', [])",
"    if current_step_index >= len(steps):",
"        raise IndexError(f'current_step_index {current_step_index} out of bounds for {len(steps)} steps')",
"    ",
"    current_step = steps[current_step_index]",
"    substeps = current_step.get('substeps', [])",
"    ",
"    # Find assistant substep content",
"    assistant_content = None",
"    for ss in substeps:",
"        if ss.get('role') == 'assistant':",
"            assistant_content = ss.get('content')",
"            break",
"    ",
"    if assistant_content is None:",
"        raise KeyError('Assistant content at current step is missing.')",
"    print(f'Assistant content at step {current_step_index}: {assistant_content}')",
"    ",
"    # Try to extract a number near the word 'available'",
"    match = re.search(r'(\\d+)\\s+(?:available)', assistant_content, flags=re.IGNORECASE)",
"    if match:",
"        stated_count = int(match.group(1))",
"    else:",
"        # Fallback: extract the first integer in the content",
"        any_num = re.search(r'(\\d+)', assistant_content)",
"        if not any_num:",
"            raise ValueError('No numeric count found in assistant content to verify.')",
"        stated_count = int(any_num.group(1))",
"    ",
"    print(f'Extracted stated_count: {stated_count}')",
"    ",
"    # Find the most recent get_product_details tool response prior to this step",
"    tool_response_content = None",
"    for idx in range(current_step_index - 1, -1, -1):",
"        step = steps[idx]",
"        for ss in step.get('substeps', []):",
"            if ss.get('role') == 'tool':",
"                # Check if previous assistant step called get_product_details",
"                # Look at the assistant step before this tool response",
"                if idx > 0:",
"                    prev_step = steps[idx - 1]",
"                    for prev_ss in prev_step.get('substeps', []):",
"                        if prev_ss.get('role') == 'assistant':",
"                            try:",
"                                calls = json.loads(prev_ss.get('content', ''))",
"                                if isinstance(calls, list):",
"                                    for call in calls:",
"                                        if call.get('function', {}).get('name') == 'get_product_details':",
"                                            tool_response_content = ss.get('content')",
"                                            print(f'Found get_product_details tool response at step {idx}')",
"                                            break",
"                            except json.JSONDecodeError:",
"                                pass",
"                if tool_response_content:",
"                    break",
"        if tool_response_content:",
"            break",
"    ",
"    if tool_response_content is None:",
"        raise KeyError('No prior get_product_details tool response found for verification.')",
"    ",
"    print(f'Raw tool response content: {tool_response_content}')",
"    ",
"    # Parse JSON content",
"    try:",
"        tool_response = json.loads(tool_response_content)",
"    except Exception as e:",
"        raise ValueError(f'Tool response content is not valid JSON: {e}')",
"    ",
"    variants = tool_response.get('variants')",
"    if variants is None:",
"        raise KeyError('Missing variants in get_product_details tool response.')",
"    if not isinstance(variants, dict):",
"        raise TypeError('variants should be a dict.')",
"    ",
"    # Compute available count",
"    available_count = 0",
"    for k, v in variants.items():",
"        if not isinstance(v, dict):",
"            print(f'Skipping variant {k}: not a dict')",
"            continue",
"        avail_flag = v.get('available', None)",
"        print(f'Variant {k} available flag: {avail_flag}')",
"        if avail_flag is True:",
"            available_count += 1",
"    ",
"    print(f'Computed available_count from tool response: {available_count}')",
"    ",
"    result = (stated_count == available_count)",
"    print(f'Assertion result (stated == computed): {result}')",
"    return result"
    ],
"nl_check": {}
}
    
\end{minted}
\section{Judge Prompts}

\subsection{Baseline Judge Prompt}
\label{app:baselinejudge_prompt}
\begin{promptbox}
You are an Expert Failure-Categorization Judge. You will be provided with a trajectory of an agent's interaction with a user.
\textbf{Given:} a full trajectory of an agent's conversation with the user (step-indexed)

\textbf{YOUR TASK:} is determine why the agent failed, which failure category applies from the taxonomy below and exactly which step index the failure occurred at. The failure taxonomy has the following categories:

\begin{enumerate}
  \item \textbf{ Instruction/Plan Adherence Failure} The agent fails to follow the directions or the agreed plan by ignoring directives and skipping policy steps.
This covers both under-execution (missed steps) and over-execution (unplanned or unnecessary actions, e.g., extra tool calls) that deviate from the static plan,
domain policy or orchestrator plan.
\item \textbf{Invention of New Information:} The agent introduces, removes, or alters information that is not grounded in any available input, context, or tool output. This includes fabricating unsupported facts, hallucinating details, or omitting relevant information without justification. 
\item \textbf{Invalid Invocation:} The agent encounters errors triggered by inputs that cannot be parsed or validated e.g., Kusto syntax errors or tool calls with bad/missing arguments. Not involving wrong logic; just invalid inputs.
  \item \textbf{Misinterpretation of Tool Output / Handoff Failure:} The agent incorrectly reasons about its own or another agent's tool output (like computation errors), leading to incorrect assumptions or actions. This also includes cases where the agent considered only partial tool output.
  \item \textbf{Intent-Plan Misalignment:} The agent misreads the user's goal or constraints and produces the wrong step sequence or structure. Covers both bad ordering/structure and plans aimed at the wrong objective.
  \item \textbf{Underspecified User Intent:} The agent was unable to complete the task due to lack of complete information at any point in the trajectory/plan execution.
  \item \textbf{Intent Not Supported:} The agent/user is asking to perform an action for which a tool is not available, like listening to an audio file.
  \item \textbf{Guardrails Triggered:} The agent is blocked by safety policies or by external site access restrictions, preventing execution despite a valid plan. Examples include policy refusals (e.g., unsafe content, privacy-protected data), CAPTCHA/robot blocks, login/paywall/403/robots.txt denials, or site forbids automation. This is not an agent planning/execution error; it is an external/guardrail block.
  \item \textbf{System Failure:} The agent faces a system connectivity issue while calling a particular tool like an endpoint not being reachable
  \item \textbf{Inconclusive (USE SPARINGLY)}: If you are not able to classify the failure into any of the above categories, label it as inconclusive and create your own category.
\end{enumerate}

\textbf{How to Judge (Decision Procedure):}
\begin{enumerate}
  \item \textbf{Step 1 - Locate the first failure:} Scan the trajectory step-by-step from the start. The first step where the agent deviates from the intended plan or emits an error is the first failure. Record the step index and a short failure note.
  \item \textbf{Step 2 - Check if that failure was resolved}: Look ahead in the trajectory for evidence that the error was resolved. If yes → Resolved; if no such evidence → Not resolved.
  \item \textbf{Step 3 - Decide and continue:}
  \begin{itemize}
    \item If Resolved: continue scanning from the next step to find the next new failure, then repeat Step 2 for it.
    \item If Not: treat this step as the root-cause failure for the run and assign the taxonomy at this step.
  \end{itemize}
\end{enumerate}

\textbf{Output format (JSON):}
\begin{minted}{json}
{
  "reason_for_failure": "string",
  "failure_case": "int 1-10",
  "reason_for_index": "string",
  "index": "int"
}
\end{minted}
\end{promptbox}

\subsection{Checklist Baseline Judge Prompt}
\label{app:checklistbaselinejudge_prompt}

\begin{promptbox}
You are an Expert Failure-Categorization Judge. You will be provided with a trajectory of an agent's interaction with a user.
\textbf{Given:} a full trajectory of an agent's conversation with the user (step-indexed)

\textbf{YOUR TASK:} is determine why the agent failed, which failure category applies from the taxonomy below and exactly which step index the failure occurred at. The failure taxonomy has the following categories:

\begin{enumerate}

\item \textbf{Instruction/Plan Adherence Failure}: Goal is correct, but the agent deviates from the required plan by ignoring directives and skipping steps despite having enough information.
This covers both under-execution (missed steps) and over-execution (unplanned or unnecessary actions, e.g., extra tool calls) that deviate from the static plan, domain policy or orchestrator plan.
Checklist:

- Can you state the user's goal, and do the agent's intent and end goal match that goal (i.e., the agent is not solving the wrong problem)?

- Was all the required information already available at this step (user intent, required context, prior tool outputs)?

- Is there a step where the ground-truth/policy requires an action (tool call, question, confirmation, ordering) and the agent did something different (skipped it / reordered it / added extra unneeded action)?

\item \textbf{Invention of New Information}: The agent introduces, removes, or alters information that is not grounded in any available input, context, or tool output. This includes fabricating unsupported facts, hallucinating details, or omitting relevant information.
Checklist:
   
- Can you pinpoint the exact invented/altered/omitted claim, value, or assumption the agent used?

- Is that claim absent from all evidence available up to that step (user text, provided context, tool outputs)?

- Did the agent rely on that claim to decide an action or produce the failing conclusion (not just harmless wording)?

\item \textbf{Invalid Invocation}: Tool call fails because the request is ill-formed (missing args, wrong fields/types, malformed query, schema mismatch).
Checklist:

- At the failure step, did the agent attempt a tool call with a concrete invocation payload/arguments?

- Does the tool/runtime explicitly report a parse/validation/schema/syntax error for that call (e.g., missing field, invalid type, cannot parse, malformed query)?

- Is the error NOT a network/timeout/service-unavailable/endpoint-unreachable issue (infra/connectivity)?

- Is the error NOT primarily a CAPTCHA/login/paywall refusal (access/guardrail block)?

\item \textbf{Misinterpretation of Tool Output / Handoff Failure}: The agent incorrectly reasons about its own or another agent's tool output, leading to incorrect assumptions or actions. This also includes cases where the agent considered only partial tool output.
Checklist:

- Before (or at) the failure step, did the agent receive tool output or handoff output that is relevant to the failing decision?

- Did the agent state or imply a specific reasoning derived from that tool output?

- Does that reasoning contradict the tool output, omit a crucial part, or reflect a clear computation/logic error relative to the output?

\item \textbf{Intent-Plan Misalignment}: Agent misunderstands the user's intent/constraints and pursues the wrong objective or violates key constraints due to misunderstanding.
Checklist:

- Do the agent's actions/plan optimize for a different goal OR violate a key constraint (not a minor wording/format issue)?

- Is the misalignment due to misunderstanding of intent/constraints (rather than missing required info from the user/context/tool outputs)?

- Is the misalignment not primarily caused by a tool error (invalid invocation, infra failure, or access/guardrail block)?

\item \textbf{Underspecified User Intent}: The agent was unable to complete the task due to lack of complete information at any point in the trajectory/plan execution.
Checklist:

- Can you identify a specific missing piece of information that is required to proceed correctly (e.g., date, address, account id, item variant)?

- Is that information absent from all evidence available up to that step (user text, provided context, and tool outputs)?

- Did the agent fail because it proceeded without obtaining this information OR because it did not ask for it when needed?

\item \textbf{Intent Not Supported}: Requested action cannot be performed with available tools/capabilities.
Checklist:
   
- Is the user requesting an action that requires an external capability/tool (e.g., listen to audio, access a private system, perform a human action)?
   
- Given the tool set available in this environment, is there no tool that can perform the requested action?
   
- Is the failure not primarily caused by infrastructure/connectivity issues?

\item \textbf{Guardrails Triggered}: The agent is blocked by safety/RAI policies or by external site access restrictions, preventing execution despite a valid plan.
Checklist:
- Is there an explicit refusal/block signal (policy refusal, CAPTCHA, login required, 403, paywall, robots.txt, automation forbidden)?

- Would the plan be feasible and correct if this block were removed (i.e., the agent is not pursuing the wrong goal/constraints)?
   
- Is the failure not primarily due to malformed tool invocation (schema/syntax/args validation error)?
   
- Is the failure not primarily due to infrastructure/connectivity issues (timeouts, endpoint unreachable)?

\item \textbf{System Failure}: The agent faces a system connectivity issue while calling a particular tool like an endpoint not being reachable.
Checklist:

- At the failure step, did the agent attempt a tool call or rely on a tool that should have been callable?

- Is there an explicit infra/connectivity error signal (timeout, connection refused, DNS failure, endpoint unreachable, service unavailable, premature termination)?

- Is the failure not primarily a parse/validation/schema/syntax error caused by malformed arguments?

\item \textbf{Inconclusive (USE SPARINGLY)}: None of 1-10 clearly apply; must provide a custom category label.
Checklist:

- If labeling as 10, did you provide a non-empty custom category describing the failure type?

\end{enumerate}
\textbf{How to Judge (Decision Procedure):}
\begin{enumerate}
  \item \textbf{Step 1 - Locate the first failure:} Scan the trajectory step-by-step from the start. The first step where the agent deviates from the intended plan or emits an error is the first failure. Record the step index and a short failure note.
  \item \textbf{Step 2 - Check if that failure was resolved}: Look ahead in the trajectory for evidence that the error was resolved. If yes → Resolved; if no such evidence → Not resolved.
  \item \textbf{Step 3 - Decide and continue:}
  \begin{itemize}
    \item If Resolved: continue scanning from the next step to find the next new failure, then repeat Step 2 for it.
    \item If Not: treat this step as the root-cause failure for the run and assign the taxonomy at this step.
  \end{itemize}
\end{enumerate}

\textbf{Output format (JSON):}
\begin{minted}{json}
{
  "taxonomy_checklist_reasoning": "string",
  "reason_for_failure": "string",
  "failure_case": "int 1-10",
  "reason_for_index": "string",
  "index": "int"
}
\end{minted}
\end{promptbox}

\subsection{\tool Judge Prompt}
\label{app:ourjudge_prompt}

\begin{promptbox}
    You are an Expert Failure-Categorization Judge. You will be provided with a trajectory of an agent's interaction with a user.
\textbf{Given:} a full trajectory of an agent's conversation with the user (step-indexed)

\textbf{YOUR TASK:} is determine why the agent failed, which failure category applies from the taxonomy below and exactly which step index the failure occurred at. The failure taxonomy has the following categories:

\begin{enumerate}
  \item \textbf{ Instruction/Plan Adherence Failure} The agent fails to follow the directions or the agreed plan by ignoring directives and skipping policy steps.
This covers both under-execution (missed steps) and over-execution (unplanned or unnecessary actions, e.g., extra tool calls) that deviate from the static plan,
domain policy or orchestrator plan.
\item \textbf{Invention of New Information:} The agent introduces, removes, or alters information that is not grounded in any available input, context, or tool output. This includes fabricating unsupported facts, hallucinating details, or omitting relevant information without justification. 
\item \textbf{Invalid Invocation:} The agent encounters errors triggered by inputs that cannot be parsed or validated e.g., Kusto syntax errors or tool calls with bad/missing arguments. Not involving wrong logic; just invalid inputs.
  \item \textbf{Misinterpretation of Tool Output / Handoff Failure:} The agent incorrectly reasons about its own or another agent's tool output (like computation errors), leading to incorrect assumptions or actions. This also includes cases where the agent considered only partial tool output.
  \item \textbf{Intent-Plan Misalignment:} The agent misreads the user's goal or constraints and produces the wrong step sequence or structure. Covers both bad ordering/structure and plans aimed at the wrong objective.
  \item \textbf{Underspecified User Intent:} The agent was unable to complete the task due to lack of complete information at any point in the trajectory/plan execution.
  \item \textbf{Intent Not Supported:} The agent/user is asking to perform an action for which a tool is not available, like listening to an audio file.
  \item \textbf{Guardrails Triggered:} The agent is blocked by safety policies or by external site access restrictions, preventing execution despite a valid plan. Examples include policy refusals (e.g., unsafe content, privacy-protected data), CAPTCHA/robot blocks, login/paywall/403/robots.txt denials, or site forbids automation. This is not an agent planning/execution error; it is an external/guardrail block.
  \item \textbf{System Failure:} The agent faces a system connectivity issue while calling a particular tool like an endpoint not being reachable
  \item \textbf{Inconclusive (USE SPARINGLY)}: If you are not able to classify the failure into any of the above categories, label it as inconclusive and create your own category.
\end{enumerate}

You are also provided a list of violations that have been generated through the trajectory through various constraints. Use these to help you identify the root cause category, failure step and agent. Static constraints have been generated through the domain policy and system prompt. Each static constraint is associated with a tool call to ensure it adheres to the domain policy. Dynamic constraints have been generated to cover computation checks, data accuracy, argument validity, and tool output consistency.
Each constraints returns a boolean, and if it returns false, it indicates a violation. Note that some violations may be false positives and not all violations may be relevant to the root cause failure.

Here are the list of violations:

\begin{tcolorbox}[
  enhanced,
  breakable,
  colback=white,
  colframe=black!40,
  boxrule=0.4pt,
  arc=2pt,
  left=6pt,right=6pt,top=6pt,bottom=6pt,
  title=\textbf{Executable check violation with grounded evidence},
  fonttitle=\small,
  coltitle=black,
]
\footnotesize
\begin{minted}[breaklines,breakanywhere,fontsize=\footnotesize]{text}
================================
VIOLATION #1
================================

Step Index: 2
Assertion Name: kusto_invocation_requires_predefined_query_and_correct_cluster
Constraint Type: CAPABILITY
Check Type: python_check
Severity: medium

Check Hint:
----------------------------
When KustoAgent runs a query, it must be a predefined query present in the plan or prior Orchestrator instruction, and the query must be tailored to the incident's cluster (no placeholders like TODO/TBD/<CLUSTER>). Verify that a kusto code block exists earlier and that the current query's clusterName matches the cluster parsed from the incident description.
----------------------------

Evidence:
-----------------------------
Current Event:
  Role: KustoAgent
  Content:
    **Kusto Query:**
    let driftedSettingName = 'VncEndpointCandidates';
    ...
    
     semantic_query_matcher: True 

Matched Substeps:
  Sub-index: 5
  Role: KustoAgent
-----------------------------

Taxonomy Targets:
  - InvalidInvocation
  - Instruction/PlanAdherenceFailure
  - IntentPlanMisalignment

\end{minted}
\end{tcolorbox}

\textbf{How to Judge (Decision Procedure):}
\begin{enumerate}
  \item \textbf{Step 1 - Locate the first failure:} Scan the trajectory step-by-step from the start. The first step where the agent deviates from the intended plan or emits an error is the first failure. Record the step index and a short failure note.
  \item \textbf{Step 2 - Check if that failure was resolved}: Look ahead in the trajectory for evidence that the error was resolved. If yes → Resolved; if no such evidence → Not resolved.
  \item \textbf{Step 3 - Decide and continue:}
  \begin{itemize}
    \item If Resolved: continue scanning from the next step to find the next new failure, then repeat Step 2 for it.
    \item If Not: treat this step as the root-cause failure for the run and assign the taxonomy at this step.
  \end{itemize}
\end{enumerate}

\textbf{Output format (JSON):}
\begin{minted}{json}
{
  "reason_for_failure": "string",
  "failure_case": "int 1-10",
  "reason_for_index": "string",
  "index": "int"
}
\end{minted}
\end{promptbox}

\section{Open Source Licensing}

$\tau$-bench, Magentic-One, and Who\&When are released under the MIT License; MAST under CC-BY 4.0. Our benchmark is released under MIT License.

\noindent Copyright (c) 2026 [Copyright Holder]
 
 \medskip
 
 \noindent Permission is hereby granted, free of charge, to any person obtaining a copy
 of this software and associated documentation files (the ``Software''), to deal
 in the Software without restriction, including without limitation the rights
 to use, copy, modify, merge, publish, distribute, sublicense, and/or sell
 copies of the Software, and to permit persons to whom the Software is
 furnished to do so, subject to the following conditions:
 
 \medskip
 
 \noindent The above copyright notice and this permission notice shall be included in all
 copies or substantial portions of the Software.
 
 \medskip
 
 \noindent \textsc{The Software is provided ``as is'', without warranty of any kind, express or
 implied, including but not limited to the warranties of merchantability,
 fitness for a particular purpose and noninfringement. In no event shall the
 authors or copyright holders be liable for any claim, damages or other
 liability, whether in an action of contract, tort or otherwise, arising from,
 out of or in connection with the Software or the use or other dealings in the
 Software.}
\end{document}

\section{Constraint Generator Prompts}
We use a single prompt template for constraint generation across domains; the only changes are domain-specific insertions such as the tool schema and policy text. We nonetheless list the instantiated prompts for the three domains for completeness.

\subsection{$\tau$-bench Prompt}
\TODO{}

\label{app:taubench-generator}
\subsection{Flash Prompt}
\TODO{}

\label{app:flash-generator}
\subsection{Magentic-One Prompt}
\TODO{}

\label{app:magentic-generator}

\subsection{One-Shot Prompt}
\TODO{}

\mypara{Plan Adherence Failure} 
\begin{enumerate}
    \item An open-source dataset of 115 failed long-horizon trajectories across 3 agentic domains—
    including step-level annotations and failure categorization.

    \item A cross-domain failure taxonomy derived through grounded theory coding,
    capturing 9 root-cause categories that generalize across diverse multi-agent settings.

    \item \tool{}, a novel Agentic failure diagnosis framework that combines
    dynamic constraint synthesis and LLM-based adjudication using auditable
    violation logs.

    \item Extensive experiments showing \TODO{XXX}\% absolute improvement in failure localization and \TODO{XXX}\% improvement in root-causing the failure.
\end{enumerate}

\mypara{Invention of new information} The agent introduces, removes, or alters information that is not grounded in any available input, context, or tool output.
This includes fabricating unsupported facts, hallucinated details, or omitting relevant information without justification.

\mypara{Invalid Invocation} The agent issues an invalid tool call, e.g., malformed inputs, missing required arguments, or values that fail schema validation.

\mypara{Misinterpretation of Tool Output} The agent incorrectly reasons about its own or another agent's tool output, leading to incorrect assumptions or actions.

\mypara{Intent–Plan Misalignment} The agent misinterprets the user's goal or constraints and produces an incorrect plan.

\mypara{Underspecified User Intent} The agent was unable to complete the task due to lack of complete information at any point in the trajectory.

\mypara{Intent Not Supported} The agent is asked to perform an action for which a tool is not available.

\mypara{Guardrails Triggered} The agent's execution is blocked by safety policies or external access restrictions despite a feasible plan.


\begin{table*}[t]
\centering
\caption{Results are mean $\pm$ std over $n{=}3$ independent runs. All runs use both programmatic and semantic checks. Metrics: step-index accuracy, category accuracy, and average step distance}
\label{tab:ablations}
\footnotesize
\setlength{\tabcolsep}{4pt}
\renewcommand{\arraystretch}{1.10}

\definecolor{blockgray}{gray}{0.88}

\begin{tabular*}{\textwidth}{@{\extracolsep{\fill}} l|c|c c c c}
\toprule
& \multicolumn{1}{c|}{} & \multicolumn{4}{c}{\textbf{\tool Ablations}} \\
\cmidrule(lr){2-2}\cmidrule(lr){3-6}
\textbf{Metric ($n{=}3$)}
& \textbf{Baseline}
& \textbf{Violations} & \textbf{Step-then-Category} & \textbf{Taxonomy Checklist} & \textbf{Checklist + Violations} \\
\midrule

\rowcolor{blockgray}
\textbf{Tau-Bench} & \multicolumn{5}{c}{\textbf{One-shot Constraint Generation}} \\
Step-index acc. (\%) $\uparrow$  & 32.2 $\pm$ 3.2 & 47.1 $\pm$ 1.6 & \fbox{\textbf{54.0 $\pm$ 1.6}} & 32.2 $\pm$ 1.6 & 48.3    \\
Category acc. (\%) $\uparrow$    & 25.3 $\pm$ 1.6 & 37.9 $\pm$ 2.8 & \fbox{\textbf{40.2 $\pm$ 1.6}} & 25.3 $\pm$ 1.6 &  39.1 $\pm$ 1.6 \\
Avg.\ step distance $\downarrow$ & 5.7  $\pm$ 0.8 & 2.8  $\pm$ 0.3 & 2.45 $\pm$ 0.5 &  5.8 $\pm$ 0.7 & 3.0  $\pm$ 0.3 \\
\rowcolor{blockgray}
\textbf{Tau-Bench} & \multicolumn{5}{c}{\textbf{Step-by-Step Constraint Generation}} \\
Step-index acc. (\%) $\uparrow$  & 32.2 $\pm$ 3.2 & -- & -- & 32.2 $\pm$ 1.6  & -- \\
Category acc. (\%) $\uparrow$    & 25.3 $\pm$ 1.6 & -- & -- & 25.3 $\pm$ 1.6  & -- \\
Avg.\ step distance $\downarrow$ & 5.7  $\pm$ 0.8 & -- & -- & 5.8  $\pm$ 0.7 & -- \\
\midrule

\rowcolor{blockgray}
\textbf{Flash} & \multicolumn{5}{c}{\textbf{One-shot Constraint Generation}} \\
Step-index acc. (\%) $\uparrow$  & 80.9 $\pm$ 2.3 & 81.8  $\pm$ 1.3 & 70.6 $\pm$ 2.3 & \fbox{\textbf{83.3 $\pm$ 2.3}} &  80.1 $\pm$ 1.3  \\
Category acc. (\%) $\uparrow$    & 53.9 $\pm$ 3.6 & 53.9 $\pm$ 7.6 & 52.3 $\pm$ 2.3 & 57.9 $\pm$ 2.7 & \textbf{58 $\pm$ 3.6} \\
Avg.\ step distance $\downarrow$ & 0.25 $\pm$ 0.2 & 0.2 $\pm$ 0 & 0.36 &  0.18 $\pm$ 0.3 & 0.2  \\
\rowcolor{blockgray}
\textbf{Flash} & \multicolumn{5}{c}{\textbf{Step-by-Step Constraint Generation}} \\
Step-index acc. (\%) $\uparrow$  & 80.9 $\pm$ 2.3 & 76.1 $\pm$ 0 & 68.2 $\pm$ 5.4 & \textbf{83.3} $\pm$ 2.3 & 76.1 $\pm$ 2.3  \\
Category acc. (\%) $\uparrow$    & 53.9 $\pm$ 3.6 & 57.1 $\pm$ 6.2 & \textbf{59.5} $\pm$ 2.3 & 57.9 $\pm$ 2.7 & \fbox{\textbf{60.3 $\pm$ 1.3}}  \\
Avg.\ step distance $\downarrow$ & 0.25 $\pm$ 0.2 & 0.3  $\pm$ 0.3 & 0.4 & 0.2 $\pm$ 0.3 & 0.3\\
\midrule

\rowcolor{blockgray}
\textbf{Magentic} & \multicolumn{5}{c}{\textbf{One-Shot Constraint Generation}} \\
Step-index acc. (\%) $\uparrow$  & 31.8  & -- & -- &  31.8  & -- \\
Category acc. (\%) $\uparrow$    & 36.36 $\pm$ 3.6 & -- & -- & 37.12 $\pm$ 5.7 & -- \\
Avg.\ step distance $\downarrow$ & 22 $\pm$ 1 & -- & -- & 25.53 $\pm$ 1.2 & -- \\
\rowcolor{blockgray}
\textbf{Magentic} & \multicolumn{5}{c}{\textbf{Step-by-Step Constraint Generation}}\\
Step-index acc. (\%) $\uparrow$  & -- & -- & -- & -- & -- \\
Category acc. (\%) $\uparrow$    & -- & -- & -- & -- & -- \\
Avg.\ step distance $\downarrow$ & -- & -- & -- & -- & -- \\
\bottomrule
\end{tabular*}
\end{table*}

\end{document}